\documentclass{article} 
\usepackage{iclr2026_conference,times}

\usepackage{amsmath,amsfonts,bm}

\def\eqref#1{equation~\ref{#1}}

\def\1{\bm{1}}

\DeclareMathAlphabet{\mathsfit}{\encodingdefault}{\sfdefault}{m}{sl}
\SetMathAlphabet{\mathsfit}{bold}{\encodingdefault}{\sfdefault}{bx}{n}

\usepackage{algorithm}
\usepackage{algorithmic} 
\usepackage{arydshln}
\usepackage{comment}
\usepackage{float}
\usepackage{hyperref}
\usepackage{graphicx}
\usepackage{subcaption}
\usepackage{url}
\usepackage{enumitem}
\usepackage{cleveref}
\usepackage{wrapfig}
\usepackage{tabularx}
\usepackage{booktabs}
\usepackage{pdflscape}
\usepackage[most]{tcolorbox}
\usepackage{fvextra}

\usepackage{xcolor}
\definecolor{darkgreen}{RGB}{0,100,0}

\usepackage{fvextra}
\tcbuselibrary{breakable}
\DefineVerbatimEnvironment{Verbatim}{Verbatim}{
  breaklines,
  breakanywhere,
  breaksymbolleft={},
  breaksymbolright={},
  breaksymbolsepleft=0pt,
  breaksymbolsepright=0pt,
  fontsize=\small,
}

\newcommand{\loose}{\looseness=-1}

\title{RoboReward: General-Purpose Vision-Language Reward Models for Robotics}

\author{Tony Lee\thanks{Core contributors. Correspondence to \texttt{tonyhlee@stanford.edu}, \texttt{ajwagen@berkeley.edu}, \texttt{pertsch@berkeley.edu}.} \\
Stanford University \\
\And
Andrew Wagenmaker\footnotemark[1] \\
UC Berkeley \\
\And
Karl Pertsch\footnotemark[1] \\
UC Berkeley \\
Stanford University \\
\AND
Percy Liang \\
Stanford University \\
\And
Sergey Levine \\
UC Berkeley \\
\And
Chelsea Finn \\
Stanford University \\
}

\newcommand{\numvlms}{22}

\newcommand{\numembodiments}{14}
\newcommand{\numtotalexamples}{54,135}
\newcommand{\numtrainexamples}{45,072}
\newcommand{\numvalexamples}{6,232}
\newcommand{\numtestexamples}{2,831}

\newcommand{\numtrainexamplesrounded}{45,000}

\newcommand{\numtestexamplesrounded}{2,800}

\iclrfinalcopy 

\begin{document}

\maketitle

\begin{abstract}
A well-designed reward is critical for effective reinforcement learning-based policy improvement. In real-world robotics, obtaining such rewards typically requires either labor-intensive human labeling or brittle, handcrafted objectives. Vision-language models (VLMs) have shown promise as automatic reward models, yet their effectiveness on real robot tasks is poorly understood. In this work, we aim to close this gap by introducing (1) \textbf{RoboReward}, a robotics reward dataset and benchmark built on large-scale real-robot corpora from Open X-Embodiment (OXE) and RoboArena, and (2) vision-language reward models trained on this dataset (RoboReward 4B/8B). Because OXE is success-heavy and lacks failure examples, we propose a \emph{negative examples data augmentation} pipeline that generates calibrated \emph{negatives} and \emph{near-misses} via counterfactual relabeling of successful episodes and temporal clipping to create partial-progress outcomes from the same videos. Using this framework, we build a large training and evaluation dataset spanning diverse tasks and embodiments to test whether state-of-the-art VLMs can reliably provide rewards for robot learning. Our evaluation of open and proprietary VLMs finds that no model excels across tasks, highlighting substantial room for improvement. We then train general-purpose 4B- and 8B-parameter models that outperform much larger VLMs in assigning rewards for short-horizon robotic tasks. Finally, we deploy the 8B model in real-robot reinforcement learning and find that it improves policy learning over Gemini Robotics-ER 1.5 while narrowing the gap to RL training with human-provided rewards. We release the full dataset, trained reward models, and evaluation suite on our website to advance the development of general-purpose reward models in robotics:  \href{https://crfm.stanford.edu/helm/robo-reward-bench}{https://crfm.stanford.edu/helm/robo-reward-bench}.
\end{abstract}

\begin{figure}[h!]
  \centering
  \includegraphics[width=\textwidth]{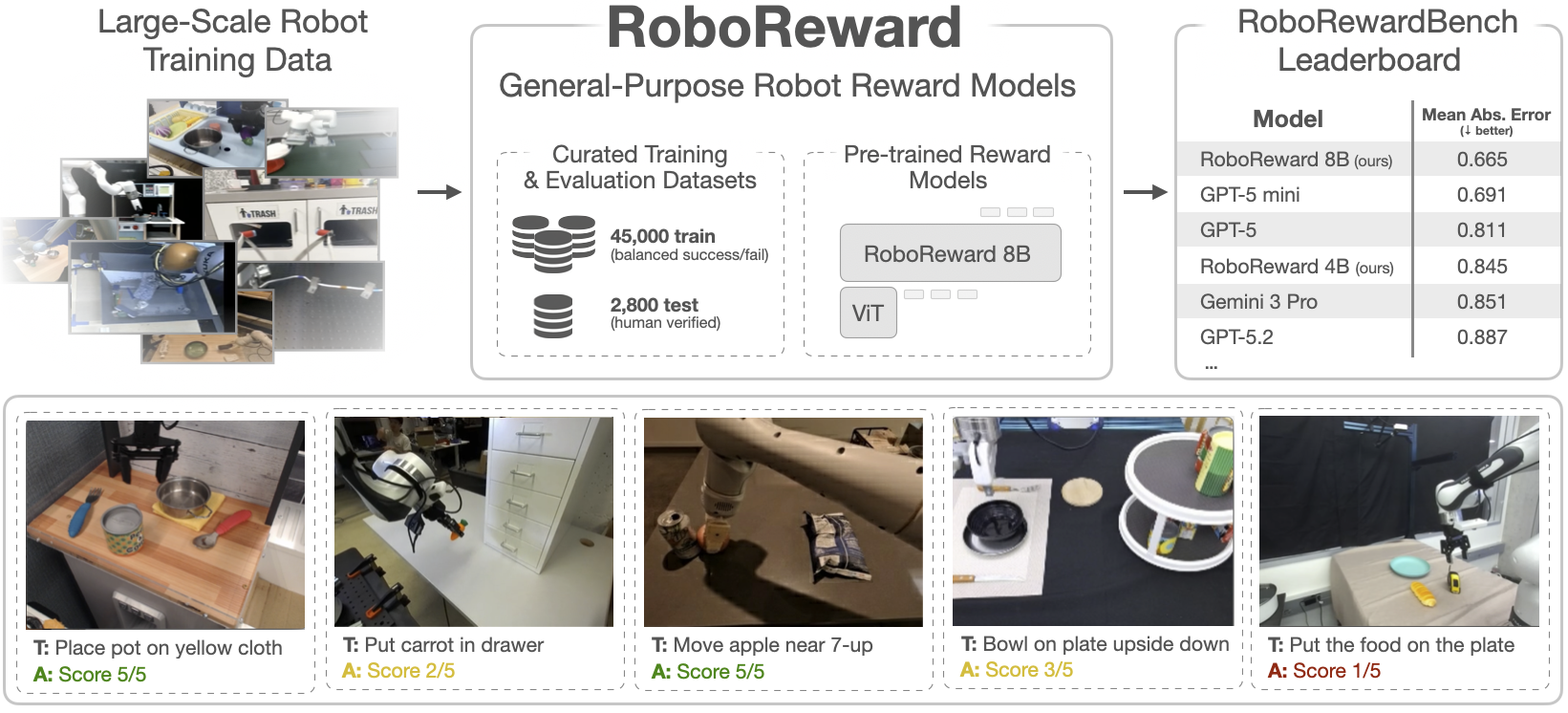}
  \caption{We introduce \textbf{RoboReward}, a dataset for training and evaluating general-purpose vision-language reward models for robotics. RoboReward consists of \numtestexamplesrounded{} real-robot episodes spanning diverse tasks and robots, with human-verified progress scores. In evaluations across \numvlms{}~proprietary and open-source VLMs, we demonstrate that today's models are lacking in their ability to provide accurate reward feedback for robots. We curate a training dataset of \numtrainexamplesrounded{} scored robot episodes across diverse embodiments and train RoboReward 4B/8B, two general-purpose vision-language reward models for robotics that outperform frontier vision-language models. We open-source all models, training data, and our evaluation benchmark to advance the development of general-purpose reward models for robotics.}
  \label{fig:teaser}
\end{figure}

\section{Introduction}

Despite recent algorithmic advances enabling efficient reinforcement learning (RL) training of robot control policies in the real world \citep{smith2022walk,luo2024serl,mark2024policy,ankile2025imitation,chen2025conrft,wagenmaker2025steering}, the broad application of RL to real-world robotics has been severely limited by the absence of accurate and informative reward models. RL-based methods critically require a precise reward signal to direct learning, yet existing methods for obtaining such rewards typically rely on either humans to label episodes by hand~\citep{myers2023active,wagenmaker2025steering}, or complex and brittle hand-crafted reward functions tuned by humans through extensive trial-and-error~\citep{lee2020learning,smith2022walk,luo2024serl,chen2025conrft}. While RL as an algorithmic paradigm holds the promise of enabling automated improvement of robot policies, the need for a human in the reward design process makes modern robotic RL labor-intensive, greatly limiting its application to general, real-world robotic policy improvement.

Motivated by these challenges, recent works have explored utilizing vision-language models (VLMs) trained on internet-scale data as automated reward models for robotics~\citep{rocamonde2023vision,venuto2024code,sontakke2024roboclip,wang2024rlvlmf}. In principle, a highly capable VLM that can reason about the physical world could replace hand-coded heuristics and expensive human supervision.
However, existing methods often fall short of achieving this, due to apparent shortcomings in current state-of-the-art VLMs and limited ability to provide sufficiently accurate rewards in real-world robot deployments. While VLMs are pretrained on large datasets drawn from a diverse set of sources---endowing them with general vision-language abilities---it is not clear that these general abilities enable them, at present, to robustly provide rewards at the level of precision and reliability required by RL training.

In this work, we seek to develop a dataset and benchmark for evaluating and improving VLM-based rewards for robotics. In simple simulation experiments, we first identify that coarse progress scores are an effective reward type for reinforcement learning, and find that reward accuracy correlates with RL performance, motivating our benchmarking design choices in a controlled setting before scaling up to a diverse, real robot dataset. Unfortunately, existing large-scale robotics datasets~\citep{open_x_embodiment_rt_x_2023, khazatsky2024droid} are heavily skewed towards successful demonstration episodes, which are poorly suited for training and evaluating reward functions for estimating both success and failure. We therefore develop a relabeling framework for synthetically augmenting demonstration data. Our framework counterfactually relabels successful episodes with failed instructions and near-miss instructions for the \emph{same} video, holding the video of the episode fixed while varying the commanded task. We additionally generate \emph{negative} and \emph{partial-progress} examples by temporally clipping successful videos to earlier endpoints, yielding calibrated near-misses from the same underlying trajectory. We use these data augmentations to construct the \textbf{RoboReward} dataset: we augment success-heavy Open X-Embodiment (OXE) episodes~\citep{openxembodiment2023} with counterfactually relabeled and temporally clipped negatives, and we additionally include RoboArena~\citep{atreya2025roboarena} as a complementary source of real-robot trajectories. Together, this yields an extensive training corpus and a human-validated evaluation dataset for reward modeling across diverse tasks and embodiments (see \Cref{fig:teaser}).

A summary of our contributions is as follows:
\begin{enumerate}[leftmargin=*]
    \item \textbf{Negative examples data augmentation.}
    We augment success-heavy robot demonstration datasets with additional \emph{wrong} and \emph{near-miss} examples via counterfactual relabeling, and by truncating successful rollouts to create partial-progress negatives. In total, this pipeline yields \numtotalexamples{} automatically generated examples across all splits, with the test set human-verified for benchmarking.
    
    \item \textbf{Robot reward benchmarking and analysis.} We analyze supervision schemes for robotic rewards, comparing binary success signals to discrete progress labels. We also show that higher-quality robot reward models lead to stronger downstream RL policies. We then introduce \textbf{RoboRewardBench}, a comprehensive and standardized evaluation of VLMs as reward models on full robot rollouts, where we assess \numvlms{}~prominent VLMs across \numtestexamples{}~robot episodes spanning diverse tasks and \numembodiments{} different types of embodiments with a mix of exocentric and egocentric camera perspectives.

    \item \textbf{Resources.} We release the \textbf{RoboReward} training dataset, \textbf{RoboRewardBench} (the human-validated evaluation benchmark derived from the test split), and \textbf{trained reward-model checkpoints} (\textbf{RoboReward 4B} and \textbf{RoboReward 8B}) that outperform larger VLMs on assigning rewards for short-horizon tasks, along with an \textbf{evaluation suite} (including a leaderboard, prompts, raw generations, and results) to advance general-purpose reward modeling in robotics.
\end{enumerate}

Our evaluation results indicate that current general-purpose VLMs are not yet reliable reward models in all robotic control settings and that the RoboReward dataset can significantly improve reward accuracy over strong general-purpose VLM baselines. Notably, our generalist 4B and 8B vision-language reward models rank \textbf{1} and \textbf{4} out of \textbf{\numvlms{}} on \textbf{RoboRewardBench}, outperforming substantially larger VLMs (including state-of-the-art proprietary models), and when used for real-world robotic RL, the \textbf{8B} model achieves higher task success than \textbf{Gemini Robotics-ER 1.5} and \textbf{substantially narrows the gap} to \textbf{human-provided rewards}.

\section{Related Work}

\textbf{Real-robot reinforcement learning.} 
Autonomously learning and improving robotic control policies through reinforcement learning is a longstanding goal in the robotics community. Despite limited early success applying RL directly in the real world \citep{riedmiller2009reinforcement,levine2016end,levine2018learning}, the majority of early work in this direction focused on learning in simulation and transferring the learned policy to the real world in deployment \citep{cutler2014reinforcement,rajeswaran2016epopt,tobin2017domain,peng2018sim,chebotar2019closing,lee2020learning,kumar2021rma}.
More recently, significant progress has been made applying RL to real-world locomotion \citep{smith2022walk,smith2022legged} and manipulation \citep{zhu2020ingredients,luo2024serl,mendonca2024continuously,luo2025precise} settings. 
These works have primarily focused on learning from scratch or with a limited number of human demonstrations, yet with the advent of ``generalist'' robot policies \citep{octo_2023,kim2024openvla,black2024pi_0}, significant attention has been devoted to developing RL algorithms that utilize such generalist policies as a starting point for learning, improving their behavior through RL in real-world deployment 
\citep{zhang2024grape,mark2024policy, nakamoto2024steering, chen2025conrft, hu2025flare, ankile2025imitation, wagenmaker2025steering, dong2025matters}.
All of these works, however, rely on either human reward supervision or hand-crafted reward functions in order to provide a signal for learning. This has greatly limited the application of RL to general robot learning settings, a challenge we aim to resolve in this work.

\textbf{Learned reward models for robotics.} To overcome the limitations of manually specified robot rewards, there is a long line of work for \emph{learning} robot reward functions. Early works learned robot rewards from human videos~\citep{sermanet2016unsupervised, shao2020concept, chen2021learning} or robot trajectories~\citep{ma2022vip, ma2023liv, yang2023robot, sontakke2024roboclip}. More recent works leverage the expressivity and common sense of VLMs to derive rewards for control. Preference-based approaches query VLMs over image and trajectory comparisons or ratings to learn reward functions and train policies in simulation or the real world~\citep{wang2024rlvlmf,venkataraman2024offlinevlm,luu2025erlvlm,singh2025varp}. A complementary direction directly derives sparse or shaped rewards from individual robot videos~\citep{du2023vision, rocamonde2023vision, baumli2023vision, yang2024rank2reward, alakuijala2024videolanguagecritic, yang2024adapt2reward, venuto2024code}. \citet{ma2024gvl} uses a VLM to perform in-context value learning. 
\citet{zhang2025rewind} propose a reward relabeling scheme based on ``rewinding'' robot demonstrations, but their approach disregards the content of the demonstration and is not evaluated using modern VLM models or diverse real robots.
Other works target specific settings such as legged locomotion from videos~\citep{zeng2024video2reward}, text-to-video diffusion-based dense rewards~\citep{chen2025tevir}, autonomous driving with language-goal rewards~\citep{huang2024vlmrl}, and real-to-sim iterative keypoint rewards~\citep{patel2025realtosim}. While these works demonstrate the promise of learned reward models for robotics, they typically focus on a single reward model trained for an individual robot setup. In contrast, our work presents, to our knowledge, the first comprehensive evaluation of \numvlms{}~modern VLMs as \emph{generalist} reward functions across a wide range of robot tasks and embodiments. Additionally, we provide an approach for counterfactual data relabeling that allows us to create large-scale training datasets for generalist reward functions and significantly improve over off-the-shelf models. 

The most relevant work to our modeling setting is \citet{tan2025robodopamine}, a concurrent work that introduces a \emph{process} reward modeling approach for high-precision robotic manipulation. In contrast, our work targets general-purpose vision-language \emph{end-of-episode} reward prediction across a broad range of real-robot tasks and embodiments, and contributes a comprehensive benchmark (RoboRewardBench) for evaluating many modern VLMs under a unified progress rubric. At the time of writing, their dataset and checkpoints have not been released, so we leave evaluating the Robo-Dopamine checkpoints with RoboRewardBench to future work.

The closest to our evaluation setting is the OpenGVL~leaderboard~\citep{opengvl2025}, which evaluates VLMs as temporal value estimators on expert videos using a Value-Order Correlation metric. OpenGVL defines only six tasks and reports results for 14 VLMs using only successful demonstration examples, however. In contrast, our work evaluates \numvlms{}~VLMs, measuring their ability to predict rewards (rather than values) on a range of successful \emph{and unsuccessful} trajectories, across diverse tasks and embodiments. We also release the prompts with videos and raw model predictions alongside our leaderboard for full transparency.

\textbf{Non-robot reward models.} With the recent success of RL approaches for post-training large language models~\citep{shao2024deepseekmathpushinglimitsmathematical,deepseekai2025deepseekr1incentivizingreasoningcapability}, there has been a large number of works on training effective reward models for LLM post-training and RL ~\citep{lightman2023letsverifystepstep, luo2025wizardmathempoweringmathematicalreasoning}. Additionally, a number of benchmarks have been proposed to evaluate these language reward models. For example, RewardBench~\citep{lambert2024rewardbench} and RewardBench~2~\citep{malik2025rewardbench2} test reward model accuracy, bias, and correlation with downstream LLM-RL performance. For multimodal settings, VLRewardBench~\citep{li2024vlrewardbench} and Multimodal~RewardBench~\citep{yasunaga2025mmrewardbench} probe VLM reward models across perception, hallucination, reasoning, safety, and preference judgments. In contrast to these works, our focus is on reward functions for \emph{robotic} tasks. As our evaluations show, the capabilities of current VLMs to adequately reward robot task performance lag far behind image or text domains, motivating our RoboReward benchmark.

\section{The Role of Reward in Reinforcement Learning}\label{sec:rl_role}

Reinforcement learning aims to find a policy $\pi$---a mapping from states to actions---that maximizes some reward $r$, typically a function of state and action. Formally, RL aims to find a policy $\pi$ with maximum \emph{expected} reward: $V^\pi := \mathbb{E}^\pi[\sum_{t \ge 0} \gamma^t r_t]$, where $\gamma \in [0,1)$ denotes a discount factor, and $r_t$ is the reward at step $t$. In robotics, we typically have some \emph{actual} objective we want to accomplish, and the reward function must be specified such that the policy learned by RL---the policy maximizing $V^\pi$---correctly achieves the desired objective. 

Our goal is to design a dataset for training and evaluating learned \emph{generalist} reward functions in robotics. The first step is to choose a reward function \emph{type} for our evaluation. For the purposes of this work, we restrict our investigation to \emph{episodic} rewards, which assign a reward value to a full episode rather than each individual step, and have become the standard choice of reward in many applications of RL to robotics \citep{luo2024serl,mark2024policy,ankile2025imitation,chen2025conrft,wagenmaker2025steering}.
Still, many design choices remain: episodic rewards can be binary or multi-valued, discrete or continuous. To guide the design of our \textbf{RoboReward} dataset, we first investigate how the choice of reward formulation affects downstream RL performance in simulated RL tasks. Concretely, we use the \texttt{Robomimic} benchmark~\citep{robomimic2021}, a simulation suite that includes several robotic manipulation tasks. We seek to understand (a) what type of reward leads to RL training that quickly learns the desired tasks and (b) what is the correlation between the \emph{accuracy} of a learned reward model and the online RL performance.
In all experiments, we utilize DSRL~\citep{wagenmaker2025steering}---a state-of-the-art RL fine-tuning algorithm---as our RL algorithm and apply it to finetune a diffusion policy pretrained on a dataset of task demonstrations included in \texttt{Robomimic} and ground truth rewards given by the simulation environment.

\begin{figure}[t]
  \centering
  \includegraphics[width=\textwidth]{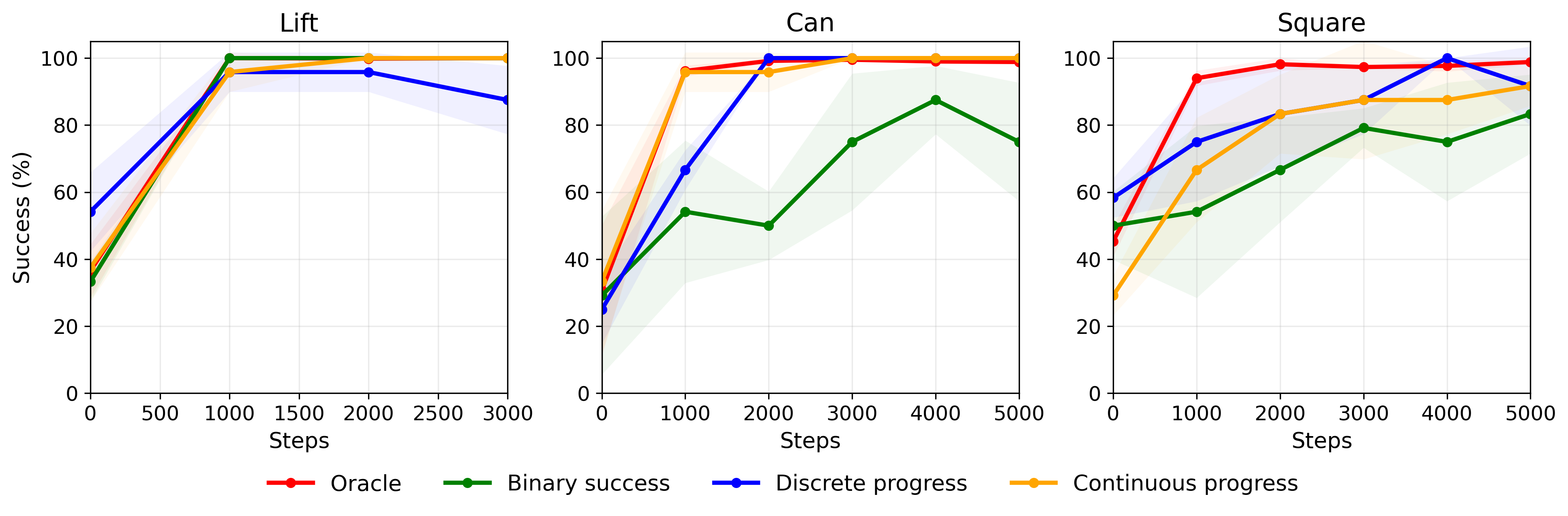}
  \caption{RL performance on three \texttt{Robomimic} tasks using learned reward functions with different reward formulations, averaged over 3 seeds (shaded regions show $\pm$1 standard deviation). Progress-based reward metrics lead to quicker convergence than a binary success metric. Both continuous and discrete progress rewards achieve comparably fast convergence. Thus, we choose \emph{discrete progress} as the reward type for our benchmark, since it leads to quick convergence and is easier for humans to annotate consistently than continuous progress.}
  \label{fig:reward_type_robomimic}
\end{figure}

\paragraph{Which reward types lead to fast RL convergence?}
We first explore what type of reward leads to the most effective RL performance. In particular, as we are primarily interested in automated, learned reward models in this work, we seek to understand what type of \emph{learned} reward leads to the most effective RL performance. We consider three different reward types:
\begin{enumerate}[leftmargin=*]
\item \textbf{Binary success}: Reward is 1 if the robot episode succeeds, and 0 otherwise.
\item \textbf{Continuous progress}: Reward is a continuous value in [0,1] corresponding to task progress given by the simulation environment.
\item \textbf{Discrete progress}: Similar to the continuous progress reward, but we discretize progress scores into 5 bins, and provide a reward in $\{1,\ldots,5\}$.
\end{enumerate}

For each reward type, we programmatically label simulated \texttt{Robomimic} episodes using the simulator’s ground-truth reward signal, and finetune a Qwen2.5-VL model~\citep{bai2025qwen25vltechnicalreport} to predict these rewards from full-episode videos.

\begin{wrapfigure}{r}{0.4\textwidth}
    \centering
    \vspace{-0.5cm}
    \includegraphics[width=\linewidth]{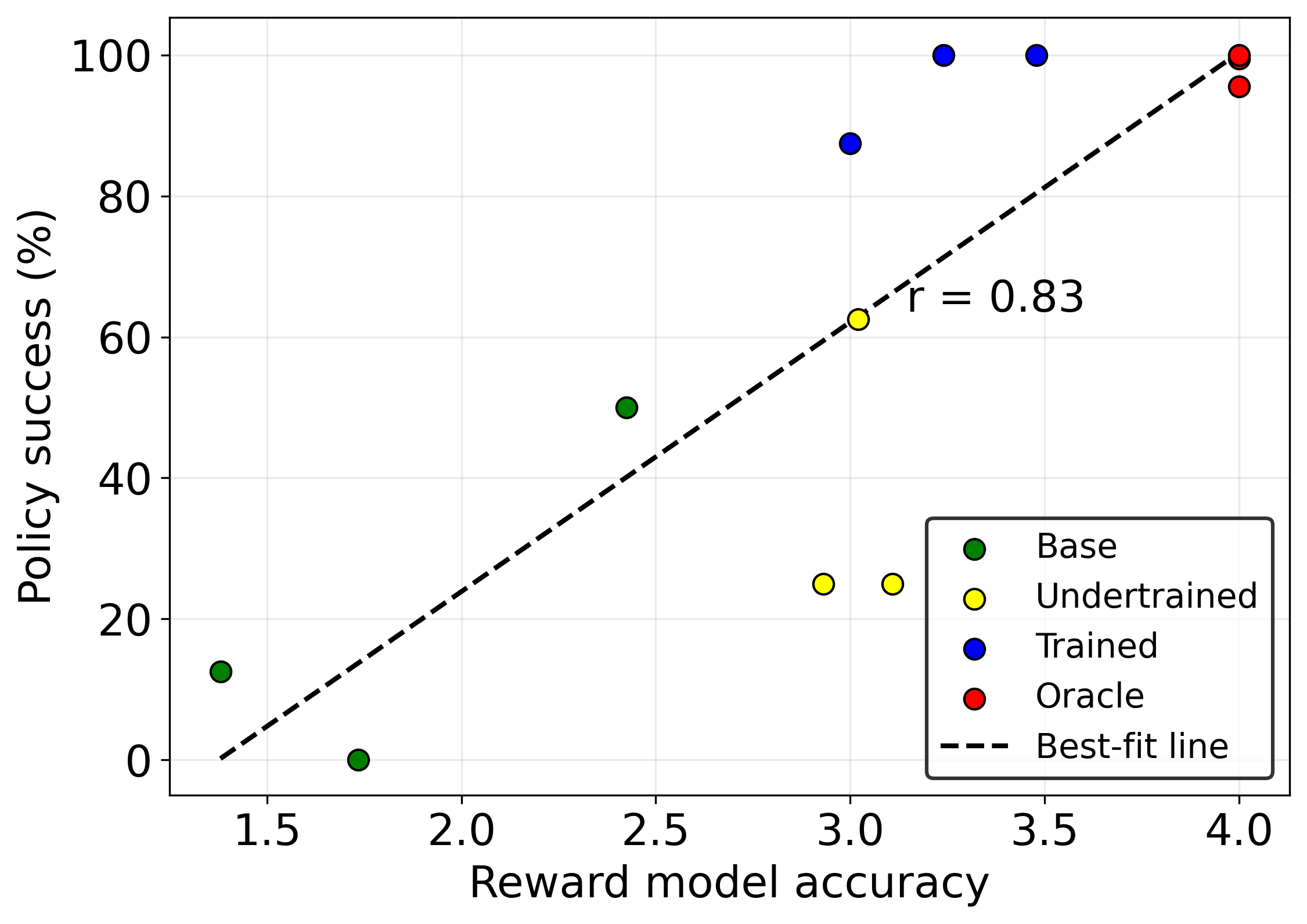}
    \caption{Strong positive correlation between reward accuracy and downstream RL performance, where the x-axis is maximum MAE minus the model MAE (larger is better; higher reward accuracy).}
    \label{fig:reward_correlation}
    \vspace{-0.5cm}
\end{wrapfigure}

The RL finetuning results are given in~\Cref{fig:reward_type_robomimic}, where we plot the true success rate against the number of samples taken. We also plot the success rate of a policy finetuned with ground-truth (binary) rewards. We observe that the type of reward significantly impacts RL performance. In particular, while both learned progress rewards perform nearly as well as the ground truth rewards, the learned binary reward performs significantly worse. This suggests that learning a progress reward for effective downstream RL performance is easier than learning a success reward and, furthermore, that whether this progress reward is discrete or continuous has minimal effect on RL performance. Thus, we choose \emph{discrete progress} as the reward formulation for RoboReward---we aim to learn a reward model that provides a progress score for a given task in $\{ 1, \ldots, 5 \}$---since it is easier for humans to annotate consistently than fully continuous rewards.

\paragraph{Do more accurate reward models lead to higher downstream RL performance?}

Next, we consider how the \emph{accuracy} of the learned reward model---in particular, reward model accuracy evaluated on held-out sets of trajectories---affects RL performance. Our primary metric throughout the paper is \textbf{mean absolute error (MAE)} between predicted and ground-truth labels (lower is better), defined as $\mathrm{MAE} := \frac{1}{N}\sum_{i=1}^{N}\left|\hat{r}_i - r_i\right|$,
where $r_i \in \{1,\dots,5\}$ is the ground-truth progress label for episode $i$ and $\hat r_i \in \{1,\dots,5\}$ is the model-predicted progress label. Intuitively, MAE measures the average distance between the predicted and true progress scores: for example, $\mathrm{MAE}<1$ means the model’s prediction is within one progress level of the ground truth on average (e.g., predicting 4 instead of 5 or 2 instead of 3).

Focusing on the discrete progress score reward from above, we measure reward accuracy on a held-out set of \texttt{Robomimic} validation episodes for multiple reward model checkpoints at different stages of convergence, as well as the off-the-shelf base model checkpoint. We then run RL to convergence with these reward models across all three \texttt{Robomimic} tasks. We show policy performance as a function of reward accuracy in \Cref{fig:reward_correlation}, where the x-axis plots the maximum possible MAE minus the model’s MAE (larger values mean higher accuracy). There is a clear correlation ($r=0.83$): more accurate rewards lead to better RL performance across the board. These results suggest that evaluating the accuracy of a reward model on a held-out offline dataset is an effective signal for determining the performance of a downstream RL application that utilizes this reward model.\loose

\section{The RoboReward Dataset and Benchmark}
\label{sec:roboreward}

To train and evaluate highly capable general reward models for robotics, we need a diverse dataset of real-world robot episodes that span successful and failed rollouts and cover a wide range of tasks and embodiments. In recent years, multiple diverse real-robot datasets have been open-sourced~\citep{open_x_embodiment_rt_x_2023, khazatsky2024droid, walke2023bridgedata, fang2024rh20t, mandlekar2018roboturk, jiang2024dexmimicgen, bharadhwaj2023roboagent, bu2025agibot}. However, most of these datasets are dominated by \emph{successful} demonstrations collected with expert policies or humans. Although this is useful for training policies with behavioral cloning, it is suboptimal for training \emph{reward models} that must discriminate fine-grained partial progress and failure. To address this imbalance, we introduce a \emph{negative-examples data augmentation} pipeline that broadens the coverage of our reward model training corpus by synthesizing \emph{partial success} and \emph{failure} from success-heavy demonstrations. This pipeline combines (1) \emph{counterfactual relabeling}: holding the video fixed while swapping in failed and near-miss instructions and (2) \emph{episode clipping}: temporally truncating successful rollouts to create partial-progress outcomes. Our approach is loosely inspired by the popular hindsight experience relabeling technique (HER, \citet{andrychowicz2017hindsight}), but instead of relabeling failed episodes as successes to increase the number of successful trials, we perform ``inverse-HER'' and relabel successes as failures to increase the number of unsuccessful trials and balance our training dataset. In this section, we describe the data sources we use to curate our RoboReward dataset, detail our relabeling procedure, and discuss the reward benchmark and the models we train on the RoboReward dataset.

\subsection{Data Sources}

We aggregate real-robot videos from two primary data sources: the Open X-Embodiment dataset (OXE, \citet{open_x_embodiment_rt_x_2023}) and RoboArena~\citep{atreya2025roboarena} evaluation data. Open X-Embodiment consists of approximately 1M real robot demonstrations, spanning 22~robot embodiments and numerous tasks, aggregated from a large number of individual academic and industry robot datasets. Each episode is paired with a natural-language instruction specifying the intended task (e.g., \textit{place the mug in the sink}), alongside the rollout video. Since many datasets in OXE are highly repetitive (most demonstrations for an individual dataset may be collected in a single scene and task setup), we uniformly subsample up to 1200 episodes per dataset to reduce overfitting. Since all OXE episodes in our dataset are demonstrations, and therefore successful examples of the given task, we assign them the maximum reward score of 5.\loose

RoboArena, on the other hand, is a diverse dataset of real-world robot policy evaluations across a broad range of scenes and tasks, using the DROID robot platform~\citep{khazatsky2024droid}. In RoboArena, human evaluators specify the natural-language tasks used for head-to-head policy evaluation and assign progress scores based on how well each policy’s rollout accomplishes the intended instruction. Since there is comparatively less repetition in RoboArena and the dataset comprises a diverse mix of successful and natural failed policy rollouts, we opt to use the full dataset without subsampling. For each episode, we map the provided human progress score (originally in range $[0, 100]$) to discrete $1, \dots, 5$ rewards. For a complete list of all RoboReward data sources and their quantities, refer to \Cref{tab:datasets}.\loose

\subsection{Data Cleaning and Negative Examples Data Augmentation}

\begin{figure}[t]
  \centering
  \includegraphics[width=\textwidth]{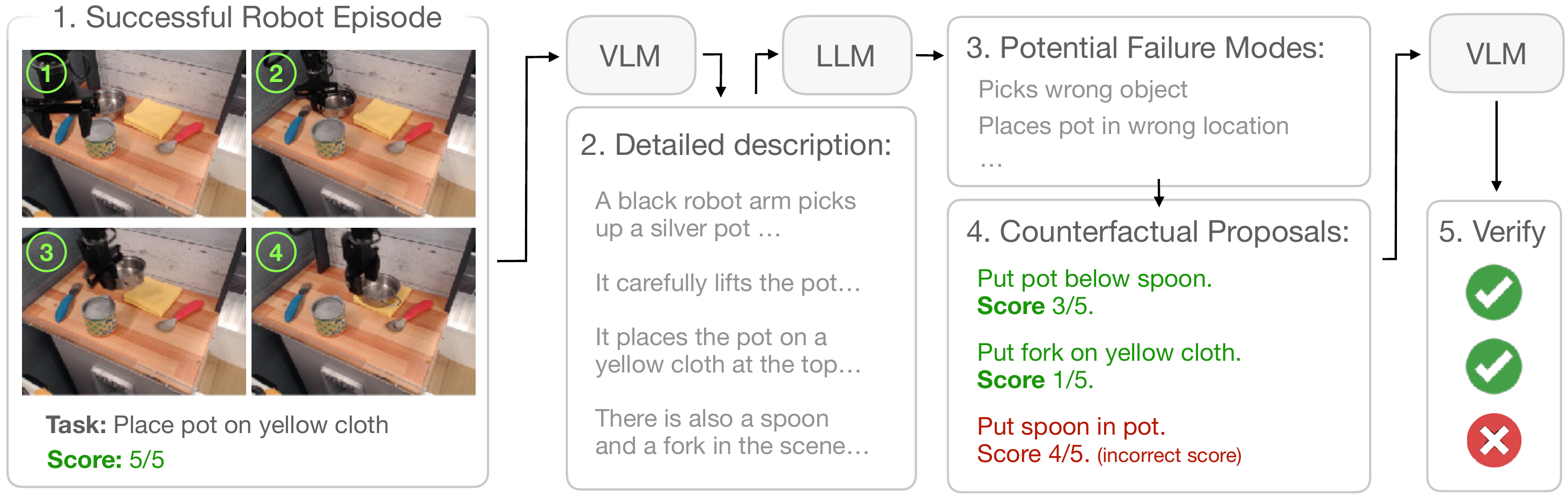}
  \caption{Overview of our counterfactual relabeling approach for generating \emph{partial success} and \emph{failure} task-video pairs for reward model training and evaluation. Given a successful robot episode, we use a VLM to describe it in detail, and then a sequence of LLM calls to propose alternative instructions for which the same video would result in only partial success or failure scores. A final VLM check verifies the quality of generated labels and rejects invalid labels.}
  \label{fig:relabeling}
\end{figure}

We now describe our data cleaning and negative examples data augmentation pipeline. Full prompts for each stage are provided in \Cref{app:cleaning}.

\paragraph{Prompt Rewriting.}
First, we normalize spelling and grammar without altering semantics, e.g., fixing spelling mistakes such as \emph{“palce dishes in the dish rack”}. We apply a text-only rewrite transform that enforces semantic invariance: it preserves the meaning while improving the surface form. We use Qwen3 Instruct (4B)~\citep{qwen3technicalreport} for this transform (see \Cref{app:prompt_rewrite}).

\paragraph{Negative Example Generation.}
Next, we address the imbalance of success vs.\ failure episodes in the data. We propose two complementary augmentations that generate additional \emph{wrong} training examples from a successful rollout video, without fabricating new videos.

\textbf{(i) Counterfactual relabeling.} Given a successful robot rollout video, we synthesize \emph{counterfactual} task commands for which the \emph{same} video would only achieve partial success, or no success at all (see \Cref{fig:relabeling}). For example, given a video of a robot placing a pepper in a pot on the stove top, our pipeline may generate alternative commands such as \texttt{place pepper in the shelf} (partial progress, since the pepper was manipulated) or \texttt{clean the pot on the stove} (no success). This yields a richer reward training dataset with a more balanced distribution of successful and failed instruction-video pairs, and encourages reward models to attend closely to the task instruction.

\textbf{(ii) Negative clipping.} In addition to modifying the task text, we create partial-progress outcomes by clipping successful videos to earlier endpoints. Concretely, for each successful episode we generate a small set of clipped videos that end at predetermined fractions of the rollout (e.g., early, mid, and late cut points), while keeping the original task instruction fixed. These clipped rollouts often exhibit minimal progress or partial completion for the original task, providing additional negative and near-miss supervision that is grounded in the same trajectory.

\paragraph{Rubric and Validation.}
We assign discrete end-state progress scores in $\{1,\dots,5\}$ using a fixed rubric:
\begin{itemize}
    \item \textit{No success} (1): The final state shows no goal-relevant change for the task command.
    \item \textit{Minimal progress} (2): The final state shows a small but insufficient change toward the goal.
    \item \textit{Partial completion} (3): The final state shows good progress but violates major requirements or multiple requirements.
    \item \textit{Near completion} (4): The final state is correct in region and intent but misses a single minor requirement.
    \item \textit{Perfect completion} (5): The final state satisfies all the requirements.
\end{itemize}

Both counterfactual relabeling and negative clipping are filtered by a VLM-based validation step: we keep only \emph{validated} examples whose task text is coherent and grounded in the video and whose provided score matches the rubric (see \Cref{app:verification} for exact prompts and additional details). 

Our label generation is intentionally an \emph{offline, high-cost} pipeline: it uses multiple stages (video analysis, rubric-grounded planning, constrained command generation, and rejection-based validation) and is therefore too slow and operationally complex to run inside an RL loop at scale. We instead use it as a \emph{teacher} to produce weak but calibrated supervision that can be distilled into a single forward-pass, open-weight reward model suitable for repeated online evaluation. While the generated labels are not guaranteed to be perfect, two factors make them useful for training: (i) we aggressively filter with a strict rubric-based validation step and discard ambiguous or inconsistent examples, which reduces label noise, and (ii) the training mixture is not purely synthetic as RoboArena provides organically occurring successes and failures with human progress scores that help ground the rewards.

\subsection{Training and Evaluation of General-Purpose Robot Reward Models}

We split the above corpus into a training and a test set. For each data source (OXE and RoboArena), we first partition by the \emph{original task descriptions} provided by the dataset, and then assign episodes to train/validation/test such that task descriptions are disjoint across splits. This mitigates train--test contamination and provides a more stringent test of generalization to unseen tasks. With the negative example data augmentation, this results in a total training set of \numtrainexamples{} episode-reward pairs, a validation set of \numvalexamples{}, and a test set of \numtestexamples{} samples.

We use the training set to finetune Qwen3-VL \citep{qwen3vl_tech_report} at two scales (4~billion and 8~billion parameters) to predict the 5-level end-of-episode progress labels when given a task description and rollout video. For both models, we freeze the vision backbone and fine-tune the fusion and LLM layers. We train for 3 epochs with cosine learning-rate decay, a warmup ratio of 0.05, weight decay of $0.05$, and max gradient norm 1.0. We use an effective batch size of 32 via gradient accumulation, with learning rates of $3\times 10^{-6}$ (4B) and $5\times 10^{-6}$ (8B). For each scale, we select the best checkpoint that minimizes the mean absolute error (MAE) between the predicted and ground-truth 1--5 reward labels on a held-out validation set, producing trained vision-language reward models: \textbf{RoboReward 4B} and \textbf{RoboReward 8B}.

We designate the \textbf{test} split as our evaluation suite. We further refine the test split by \textbf{human-verifying every example} --- the human annotator is asked to confirm that the end-state reward label is justified given the video of the rollout and task description (see \Cref{app:human_verification}). We discard any example that does not pass human verification and then subsample from the remaining verified examples to form a clean evaluation set. We refer to this human-verified test split as \textbf{RoboRewardBench}. Finally, our reported benchmark numbers are computed on \textbf{RoboRewardBench}, whose labels are \textbf{human-verified}, ensuring that evaluation remains trustworthy even if the training labels contain noise.

\section{Experiments}
\label{sec:experiments}

We next present experimental results in three parts. We begin by benchmarking off-the-shelf VLMs and our RoboReward models on RoboRewardBench (\Cref{sec:rrbench_results}). We then evaluate whether these reward models enable effective real-world policy improvement via reinforcement learning (\Cref{sec:real_world_rl}). Finally, we justify our data mixture and augmentation pipeline via data-mixture ablations that isolate the contributions of counterfactual relabeling and negative clipping to overall reward accuracy (\Cref{sec:ablations}).

\subsection{Benchmarking frontier VLMs with RoboRewardBench}\label{sec:rrbench_results}

We evaluate \numvlms{} VLMs varying in size, model developers, and access (e.g., open weights vs.\ API) on RoboRewardBench, including our trained RoboReward VLMs (see \Cref{tab:models} for the complete list). Each model is prompted with a task description and rollout video and must predict a discrete end-of-episode progress label in $\{1,2,3,4,5\}$. We measure performance using mean absolute error (MAE; lower is better), as defined in \Cref{sec:rl_role}.
Table~\ref{tab:rrbench_select} summarizes results for a select handful of representative models. For ranking the \numvlms{} models and full per-subset results, see \Cref{fig:all}, which reports the complete ranking and results for every RoboRewardBench subset.

\begin{table}[t]
\centering
\small
\setlength{\tabcolsep}{7pt}
\renewcommand{\arraystretch}{1.15}
\begin{tabular}{r l ccc}
\toprule
\textbf{Rank} & \textbf{Model} & \textbf{Overall (MAE)} $\downarrow$ & \textbf{RoboArena} $\downarrow$ & \textbf{OXE subsets} $\downarrow$ \\
\midrule
1  & RoboReward 8B \textbf{(ours)}                 & \textbf{0.665} & \textbf{0.768} & \textbf{0.660} \\
2  & GPT-5 mini (2025-08-07)         & 0.691 & 0.862 & 0.683 \\
3  & GPT-5 (2025-08-07)              & 0.811 & 1.028 & 0.801 \\
4  & RoboReward 4B \textbf{(ours)}                 & 0.845 & 0.806 & 0.847 \\
5  & Gemini 3 Pro (Preview)          & 0.851 & 1.234 & 0.833 \\
\addlinespace[2pt]
\multicolumn{5}{c}{\dots} \\
7  & Qwen3-VL Instruct (8B)          & 0.892 & 0.847 & 0.894 \\
\addlinespace[2pt]
\multicolumn{5}{c}{\dots} \\
9  & Gemini 2.5 Pro                  & 0.902 & 0.936 & 0.900 \\
\addlinespace[2pt]
\multicolumn{5}{c}{\dots} \\
11 & Gemini Robotics-ER 1.5          & 0.906 & 1.002 & 0.902 \\
\addlinespace[2pt]
\multicolumn{5}{c}{\dots} \\
13 & Gemini 2.5 Flash                & 0.943 & 1.190 & 0.931 \\
\addlinespace[2pt]
\multicolumn{5}{c}{\dots} \\
16 & Qwen3-VL Instruct (4B)          & 1.032 & 0.929 & 1.036 \\
\addlinespace[2pt]
\multicolumn{5}{c}{\dots} \\
21 & Llama 4 Scout Instruct          & 1.485 & 1.830 & 1.469 \\
22 & Qwen2.5-VL Instruct (3B)        & 1.607 & 1.443 & 1.614 \\
\bottomrule
\end{tabular}
\caption{RoboRewardBench results for a select subset of representative models. \textbf{Overall} reports the group-wise MAE (lower is better) over all RoboRewardBench subsets, \textbf{RoboArena} reports MAE on the RoboArena subset only, and \textbf{OXE subsets} reports the group-wise MAE averaged across the OXE-derived subsets. The fully expanded results and the complete ranking over all \numvlms{} models are in \Cref{fig:all}.}
\label{tab:rrbench_select}
\end{table}

\paragraph{Supervision with RoboReward yields capable general-purpose reward models.}
Our finetuned \textbf{RoboReward 8B} achieves the lowest overall MAE (0.665), outperforming all evaluated frontier VLMs, including GPT-5 mini (0.691) and Gemini family models (e.g., Gemini 2.5 Pro at 0.902, Gemini 2.5 Flash at 0.943). This trend also holds on the largest and most \emph{organic} subset, \textbf{RoboArena} (1000 episodes). Both \textbf{RoboReward 8B} (0.768) and \textbf{RoboReward 4B} (0.806) outperform all evaluated frontier VLMs, improving over strong closed-access API baselines such as GPT-5 mini (0.862), GPT-5 (1.028), and Gemini Robotics-ER 1.5 (1.002), as well as Gemini 2.5 Pro (0.936) and Gemini 2.5 Flash (1.190).

Targeted supervision provides large gains at fixed architecture: \textbf{RoboReward (8B)} improves over \textbf{Qwen3-VL Instruct (8B)} by 0.227 MAE (0.892 to 0.665) and improves on \textbf{RoboArena} from 0.847 to 0.768, and \textbf{RoboReward (4B)} improves over \textbf{Qwen3-VL Instruct (4B)} by 0.187 MAE (1.032 to 0.845) and improves on \textbf{RoboArena} from 0.929 to 0.806. Together, these results suggest that \emph{high-quality, diverse reward supervision} can substantially improve reward prediction accuracy.

\paragraph{RoboRewardBench reveals large, non-uniform generalization gaps.}
The per-subset columns in \Cref{fig:all} show pronounced performance swings across embodiments, scenes, and viewpoints. For instance, \textbf{GPT-5 (2025-08-07)} is among the top models on RoboRewardBench overall (MAE = 0.811), yet its accuracy varies widely across subsets: it is relatively strong on \textit{Austin Sirius} (0.310) and \textit{Austin VIOLA} (0.283), but substantially worse on \textit{Berkeley RPT} (1.174), \textit{Tokyo PR2 Tabletop Manipulation} (1.366), and \textit{RoboArena} (1.028). A similar non-uniformity appears for \textbf{Gemini 3 Pro (Preview)}, Google DeepMind's flagship model. Gemini 3 Pro is the best-performing model on \textit{Berkeley RPT} (0.395) and performs strongly on \textit{UCSD Pick Place} (0.37), yet it degrades sharply on other subsets such as \textit{Berkeley MVP} (1.394) and \textit{KAIST Nonprehensile Objects} (1.491). 

This generalization gap is a challenge even for models trained on robotics data. \textbf{Gemini Robotics-Embodied Reasoning 1.5} (Gemini Robotics-ER 1.5, \citet{google_gemini_robotics_er}), a frontier model trained on robotics data to excel in progress estimation and embodied reasoning, attains an overall MAE of 0.906 on RoboRewardBench, ranking 11th out of 22 models in \Cref{tab:rrbench_select}. At the same time, despite \textbf{RoboReward (8B)} ranked as the best model overall, it is not uniformly best across subsets. For example, it is comparatively weak on \textit{UTokyo xArm Bimanual} (absolute error 1.394), \textit{Stanford HYDRA} (0.890, rank 8/22), while the best model (\textbf{Gemini 2.5 Pro}) achieves an MAE of 0.495. This echoes broader findings that real-world physical reasoning remains challenging even for models trained specifically on robot data and frontier VLMs \citep{ pătrăucean2023perceptiontestdiagnosticbenchmark,lee2024vhelmholisticevaluationvision,zhang2025mmerealworldmultimodalllmchallenge}.

\subsection{Training real-robot policies with VLM reward models}\label{sec:real_world_rl}
Finally, we aim to demonstrate that RoboReward provides a sufficiently accurate reward signal to enable real-world robotic RL.

\begin{wrapfigure}{r}{0.65\textwidth}
  \centering
  \vspace{-1em} 
  \includegraphics[width=0.49\linewidth,clip,trim=0 205 0 70]{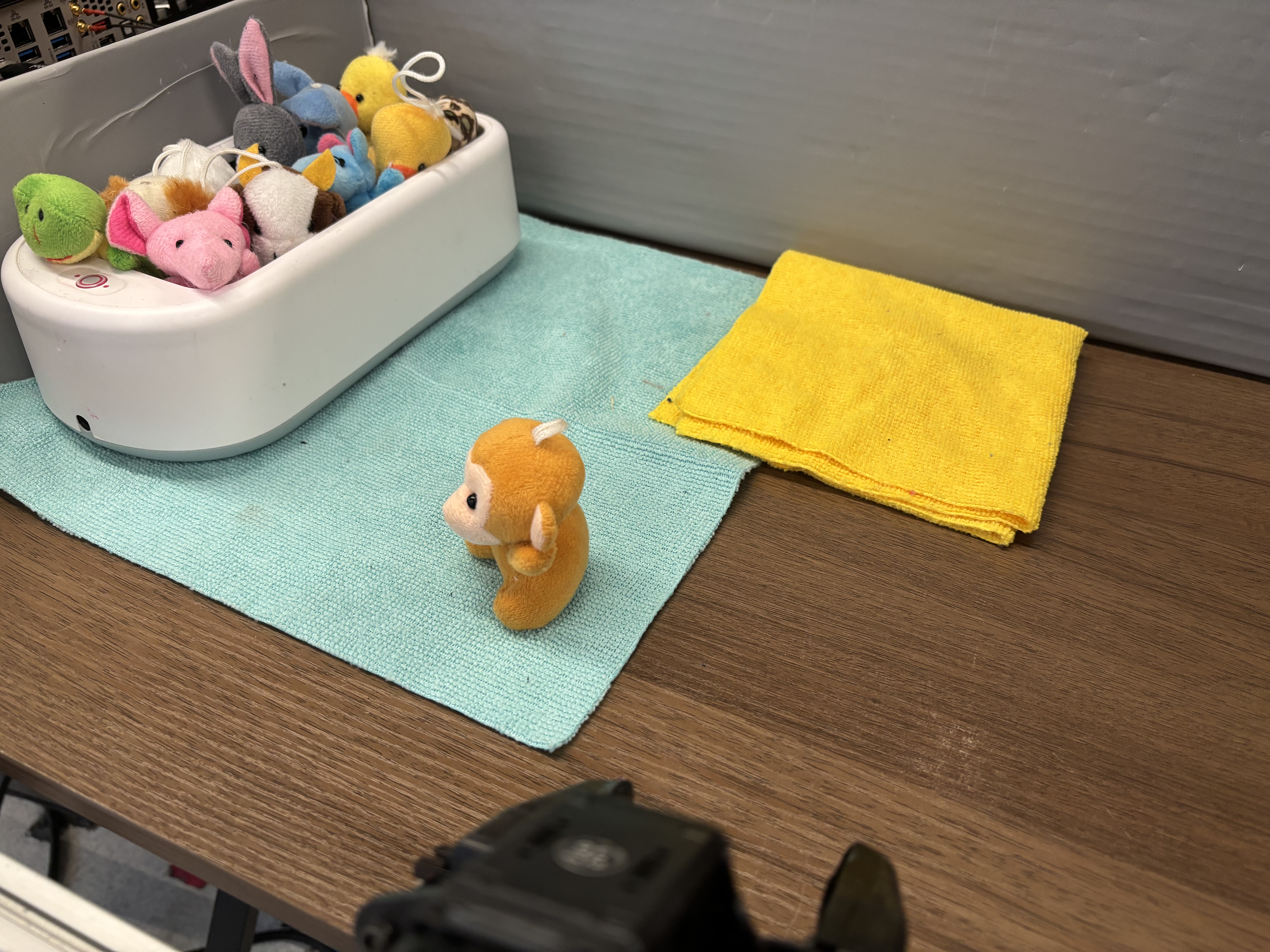}
  \includegraphics[width=0.49\linewidth,clip,trim=0 124 0 4]{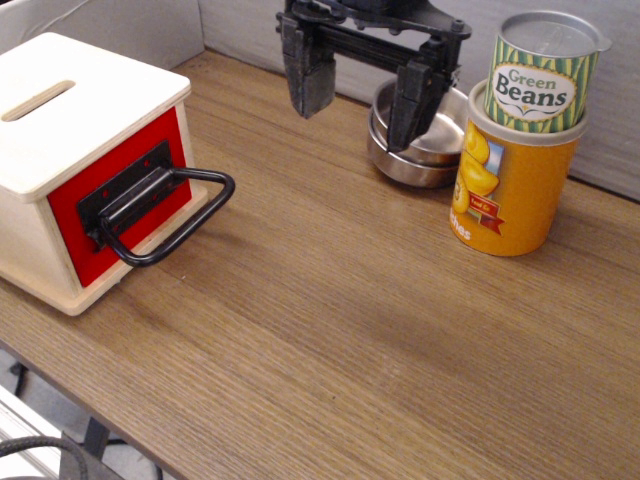}
  \caption{Real robot tasks: \emph{Pick up the brown monkey and move it on top of the yellow towel} (\texttt{Pick-and-place monkey}, \textit{left}) and \emph{Pull the drawer out} (\texttt{Open drawer}, \textit{right}). We run real-world RL improvement on each of these tasks, using RoboReward as a reward.}
  \label{fig:task_examples}
  \vspace{-1em} 
\end{wrapfigure}

\paragraph{Setting.}
For the real-world experiments, we utilize the WidowX 250 6-DoF robot arm.
As our RL algorithm, we run DSRL to finetune a multi-task diffusion policy trained on the BridgeData V2 dataset \citep{walke2023bridgedata}.
For a reward signal, we use a sparse end-of-episode reward, comparing the following three settings: (1) oracle human reward: a human labeler gives a positive reward of +1 on success and the reward is otherwise 0, (2) RoboReward 8B: outputs a 1-5 progress score at the end of each episode, and (3) Gemini Robotics-ER 1.5: outputs a progress score 1-5, similar to RoboReward. We note that Gemini Robotics-ER 1.5 is a frontier model that was specifically trained for embodied reasoning and robotics.
\textbf{Both VLM reward models are not further fine-tuned and are prompted zero-shot.}
We consider two real-world tasks for the WidowX robot to perform in two different settings, neither of which was seen during training. The first task is to pick up the brown monkey and move it on top of the yellow towel (\texttt{Pick-and-place monkey}). The second task is to pull the drawer out (\texttt{Open drawer}). See \Cref{fig:task_examples} for an illustration of each task. 
For both tasks and across the three reward settings, we train for 6000 steps, where an episode is a maximum of 70 steps (see \Cref{app:real_world_rl} for full hyperparameter and training details).

\begin{table}[!htbp]
\centering
\caption{Performance (success rate over 20 trials) of running RL with various reward models compared to the base policy. Values in parentheses show the change compared to the base policy before training with RL.  \textbf{RoboReward 8B} closes much of the gap to \textbf{human rewards}  and substantially outperforms \textbf{Gemini Robotics-ER 1.5} as a reward model.}
\label{tab:real_robot_results}
\setlength{\tabcolsep}{8pt}
\renewcommand{\arraystretch}{1.15}
\begin{tabular}{lcc}
\toprule
\textbf{Rewards} & \texttt{Pick-and-place monkey} & \texttt{Open drawer} \\
\midrule
Base policy & 5\% & 10\% \\
\addlinespace[3pt]
\cdashline{1-3}
\addlinespace[3pt]
Human & 75\% (\textcolor{darkgreen}{+70}) & 90\% (\textcolor{darkgreen}{+80}) \\
RoboReward 8B & 50\% (\textcolor{darkgreen}{+45}) & 80\% (\textcolor{darkgreen}{+70}) \\
Gemini Robotics-ER 1.5 & 10\% (\textcolor{darkgreen}{+5}) & 45\% (\textcolor{darkgreen}{+35}) \\
\bottomrule
\end{tabular}
\end{table}

\paragraph{Results.}
The results, obtained from 20 trials per task across the different settings, are summarized in \Cref{tab:real_robot_results}.
We see that RoboReward 8B enables effective real-world RL improvement, and that running RL with this reward improves the performance of the base policy on both tasks  (from $5\%$ to $50\%$ on pick-and-place monkey, and $10\%$ to $80\%$ on open drawer). While the performance of RL improvement with RoboReward does not yet match the performance achieved with the oracle human reward (which reaches $75\%$ and $90\%$, respectively, on the two tasks), it closes much of the gap, especially on \texttt{Open drawer} ($80\%$ vs.\ $90\%$). 
Furthermore, RoboReward significantly outperforms Gemini Robotics-ER 1.5, which fails to yield any improvement on \texttt{Pick-and-place monkey}, and only minimal improvement on \texttt{Open drawer}.

We also note that the ordering of models by the real-world RL performance they induce is consistent with their performance on RoboRewardBench.
In particular, Gemini Robotics-ER 1.5 has a higher overall mean absolute error on RoboRewardBench (0.906), and it underperforms RoboReward 8B as a reward model, yielding substantially weaker downstream RL gains, while RoboReward 8B has a lower mean absolute error on RoboRewardBench (0.665) and produces much greater improvements over the base policy on both tasks. These results align with our previous findings from our simulation experiments that better reward quality leads to improved downstream RL performance, and they further stress the importance of training high-quality reward models for robotics reinforcement learning. We further discuss qualitative failure modes of VLM reward models in \Cref{app:vlm_pitfalls}, which help explain the remaining gap between human-assigned rewards and VLM-predicted rewards.

\subsection{Data Mixture Ablations}\label{sec:ablations}
\begin{wraptable}{r}{0.52\textwidth}
\vspace{-1\baselineskip}
\centering
\small
\setlength{\tabcolsep}{2pt}
\renewcommand{\arraystretch}{1.15}
\begin{tabular}{lcc}
\toprule
\textbf{Data mixture} & \textbf{Overall (MAE)} $\downarrow$ & \textbf{RoboArena} $\downarrow$ \\
\midrule
Full & 0.845 & 0.806 \\
Full $-$ Neg. Aug. & 1.450 & 0.797 \\
Full $-$ Neg. clipping & 1.075 & 0.813 \\
\bottomrule
\end{tabular}
\caption{Results from data mixture ablations on RoboRewardBench (lower is better). \textbf{Overall} reports MAE on the full RoboRewardBench (including RoboArena), while \textbf{RoboArena} reports MAE on the RoboArena subset only. Training with RoboArena and OXE success examples only (without negative examples augmentation) matches in-domain errors but fails to generalize, while training on the \textbf{full RoboReward dataset} (counterfactual labels, clipped examples, and RoboArena) achieves matching performance on RoboArena and substantially improves overall performance.}
\label{tab:data_mixture_ablations}
\vspace{-1.5\baselineskip}
\end{wraptable}

We evaluate how different sources of negative supervision affect the accuracy of reward models on RoboRewardBench. Our full training set includes (1) successful OXE demonstrations augmented with counterfactual relabeling and temporally clipped versions of the same videos to create visually grounded negatives and near-misses, and (2) RoboArena rollouts on the DROID platform that contain organic successes and failures. Using the 4B model, we run two data-mixture ablations: (1) train on RoboArena plus the original (success-only) OXE demonstrations with no negative examples augmentation, and (2) remove the clipped-video negatives while keeping the counterfactual relabeling. \Cref{tab:data_mixture_ablations} summarizes the results. From these ablations, we highlight two main takeaways.\loose

\paragraph{RoboArena failures alone are insufficient for broad generalization.} 
Training only on RoboArena and the original OXE examples achieves nearly the same error on the RoboArena subset of RoboRewardBench as the training on the full data mixture (MAE 0.797 vs. 0.806). However, just training on RoboArena and OXE performs dramatically worse on the overall benchmark (1.45 vs. 0.845). This gap suggests that while RoboArena provides high-quality organic failures, it is limited in coverage (single platform) and does not expose the model to the broader distribution of instruction-conditioned failures present in RoboRewardBench. In contrast, counterfactual relabeling and clipping introduce diverse negative outcomes across many scenes and embodiments, which is necessary for a general-purpose reward model.

\paragraph{Clipping episodes improves robustness without harming organic performance.}
Removing clipped negative examples degrades overall performance (1.075 vs 0.845), while having minimal impact on RoboArena (0.813 vs 0.806), suggesting clipped examples primarily broaden coverage of failure modes and are critical to achieving effective performance across RoboRewardBench.

Therefore, high-quality organic data from RoboArena is valuable, but to achieve strong performance across RoboRewardBench, the reward model needs broad and instruction-sensitive negative coverage, which is provided by counterfactual relabeling and reinforced by negative clipping.

\section{Discussion}

In this work, we introduced the \textbf{RoboReward} training dataset, \textbf{RoboRewardBench}, a human-verified evaluation suite for benchmarking generalist vision-language reward models on real-robot rollouts, and two finetuned reward VLMs (\textbf{RoboReward 4B/8B}). Across \numvlms{} frontier and open-weight models, RoboReward 8B achieves the best overall accuracy on RoboRewardBench, demonstrating that targeted reward supervision can outperform substantially larger general-purpose VLMs. Further, we show in both simulation and real-robot settings that improvements in offline reward accuracy translate into improved downstream RL performance, making benchmark progress a meaningful proxy for real-world usefulness.

More broadly, our results suggest that reward modeling is a key bottleneck for real-world RL: current frontier VLMs still exhibit large and non-uniform generalization gaps across embodiments and scenes, and even small reward mistakes can meaningfully impact policy improvement. By releasing a standardized benchmark with human-verified labels and an open evaluation suite, we aim to make these limitations measurable and to accelerate progress on reliable, general-purpose reward models for robotics. An important direction for future work is extending reward modeling to longer-horizon, multi-stage tasks, where credit assignment and progress estimation become more challenging.

\bigskip
\subsection*{Acknowledgments}

We thank Yifan Mai for maintaining the HELM codebase, which was used for benchmarking and populating the leaderboard. This research was partly supported by ONR N00014-22-1-2621, ONR N00014-25-1-2060, and the Toyota Research Institute (TRI).


\bibliography{iclr2026_conference}

@misc{openai_gpt51_system_card,
  author       = {{OpenAI}},
  title        = {{GPT-5.1 Instant and GPT-5.1 Thinking System Card}},
  year         = {2025},
  howpublished = {\url{https://cdn.openai.com/pdf/4173ec8d-1229-47db-96de-06d87147e07e/5_1_system_card.pdf}},
  note         = {Accessed: 2025-12-23}
}

@misc{openai_gpt52_system_card,
  author       = {{OpenAI}},
  title        = {{Update to GPT-5 System Card: GPT-5.2}},
  year         = {2025},
  howpublished = {\url{https://cdn.openai.com/pdf/3a4153c8-c748-4b71-8e31-aecbde944f8d/oai_5_2_system-card.pdf}},
  note         = {Accessed: 2025-12-23}
}

@misc{google_gemini3_pro_model_card,
  author       = {{Google DeepMind}},
  title        = {{Gemini 3 Pro Model Card}},
  year         = {2025},
  howpublished = {\url{https://storage.googleapis.com/deepmind-media/Model-Cards/Gemini-3-Pro-Model-Card.pdf}},
  note         = {Accessed: 2025-12-23}
}

@misc{google_gemini3_flash_model_card,
  author       = {{Google DeepMind}},
  title        = {{Gemini 3 Flash Model Card}},
  year         = {2025},
  howpublished = {\url{https://storage.googleapis.com/deepmind-media/Model-Cards/Gemini-3-Flash-Model-Card.pdf}},
  note         = {Accessed: 2025-12-23}
}

@misc{google_gemini25_tech_report,
  author       = {{Google DeepMind}},
  title        = {{Gemini 2.5: Pushing the Frontier with Advanced Reasoning, Multimodality, Long Context, and Next Generation Agentic Capabilities}},
  year         = {2025},
  howpublished = {\url{https://storage.googleapis.com/deepmind-media/gemini/gemini_v2_5_report.pdf}},
  note         = {Accessed: 2025-12-23}
}

@misc{google_gemini25_flashlite_model_card,
  author       = {{Google DeepMind}},
  title        = {{Gemini 2.5 Flash-Lite Model Card}},
  year         = {2025},
  howpublished = {\url{https://storage.googleapis.com/deepmind-media/Model-Cards/Gemini-2-5-Flash-Lite-Model-Card.pdf}},
  note         = {Accessed: 2025-12-23}
}

@misc{google_gemini_robotics15_tech_report,
  author       = {{Google DeepMind}},
  title        = {{Gemini Robotics 1.5: Pushing the Frontier of Generalist Robotics}},
  year         = {2025},
  howpublished = {\url{https://storage.googleapis.com/deepmind-media/gemini-robotics/Gemini-Robotics-1-5-Tech-Report.pdf}},
  note         = {Accessed: 2025-12-23}
}

@misc{qwen3vl_tech_report,
      title={Qwen3-VL Technical Report}, 
      author={Shuai Bai and Yuxuan Cai and Ruizhe Chen and Keqin Chen and Xionghui Chen and Zesen Cheng and Lianghao Deng and Wei Ding and Chang Gao and Chunjiang Ge and Wenbin Ge and Zhifang Guo and Qidong Huang and Jie Huang and Fei Huang and Binyuan Hui and Shutong Jiang and Zhaohai Li and Mingsheng Li and Mei Li and Kaixin Li and Zicheng Lin and Junyang Lin and Xuejing Liu and Jiawei Liu and Chenglong Liu and Yang Liu and Dayiheng Liu and Shixuan Liu and Dunjie Lu and Ruilin Luo and Chenxu Lv and Rui Men and Lingchen Meng and Xuancheng Ren and Xingzhang Ren and Sibo Song and Yuchong Sun and Jun Tang and Jianhong Tu and Jianqiang Wan and Peng Wang and Pengfei Wang and Qiuyue Wang and Yuxuan Wang and Tianbao Xie and Yiheng Xu and Haiyang Xu and Jin Xu and Zhibo Yang and Mingkun Yang and Jianxin Yang and An Yang and Bowen Yu and Fei Zhang and Hang Zhang and Xi Zhang and Bo Zheng and Humen Zhong and Jingren Zhou and Fan Zhou and Jing Zhou and Yuanzhi Zhu and Ke Zhu},
      year={2025},
      eprint={2511.21631},
      archivePrefix={arXiv},
      primaryClass={cs.CV},
      url={https://arxiv.org/abs/2511.21631}, 
}

@misc{meta_llama4_model_card,
  author       = {{Meta}},
  title        = {{Llama 4: Model Cards and Prompt Formats}},
  year         = {2025},
  howpublished = {\url{https://www.llama.com/docs/model-cards-and-prompt-formats/llama4/}},
  note         = {Accessed: 2025-12-23}
}

@misc{openai_gpt5_system_card,
  author       = {{OpenAI}},
  title        = {{GPT-5 System Card}},
  year         = {2025},
  howpublished = {\url{https://cdn.openai.com/gpt-5-system-card.pdf}},
  note         = {Accessed: 2025-12-23}
}

@misc{google_gemini_robotics_er,
  author       = {{Google DeepMind}},
  title        = {{Gemini Robotics ER}},
  year         = {2025},
  howpublished = {\url{https://deepmind.google/models/gemini-robotics/gemini-robotics-er/}},
  note         = {Accessed: 2025-12-23}
}

@article{tan2025robodopamine,
    title={Robo-Dopamine: General Process Reward Modeling for High-Precision Robotic Manipulation}, 
    author={Tan, Huajie and Chen, Sixiang and Xu, Yijie and Wang, Zixiao and Ji, Yuheng and Chi, Cheng and Lyu, Yaoxu and Zhao, Zhongxia and Chen, Xiansheng and Co, Peterson and Xie, Shaoxuan and Yao, Guocai and Wang, Pengwei and Wang, Zhongyuan and Zhang, Shanghang},
    journal={arXiv preprint arXiv:2512.23703},
    year={2025}
}

@misc{zhang2025mmerealworldmultimodalllmchallenge,
      title={MME-RealWorld: Could Your Multimodal LLM Challenge High-Resolution Real-World Scenarios that are Difficult for Humans?}, 
      author={Yi-Fan Zhang and Huanyu Zhang and Haochen Tian and Chaoyou Fu and Shuangqing Zhang and Junfei Wu and Feng Li and Kun Wang and Qingsong Wen and Zhang Zhang and Liang Wang and Rong Jin and Tieniu Tan},
      year={2025},
      eprint={2408.13257},
      archivePrefix={arXiv},
      primaryClass={cs.CV},
      url={https://arxiv.org/abs/2408.13257}, 
}

@misc{pătrăucean2023perceptiontestdiagnosticbenchmark,
      title={Perception Test: A Diagnostic Benchmark for Multimodal Video Models}, 
      author={Viorica Pătrăucean and Lucas Smaira and Ankush Gupta and Adrià Recasens Continente and Larisa Markeeva and Dylan Banarse and Skanda Koppula and Joseph Heyward and Mateusz Malinowski and Yi Yang and Carl Doersch and Tatiana Matejovicova and Yury Sulsky and Antoine Miech and Alex Frechette and Hanna Klimczak and Raphael Koster and Junlin Zhang and Stephanie Winkler and Yusuf Aytar and Simon Osindero and Dima Damen and Andrew Zisserman and João Carreira},
      year={2023},
      eprint={2305.13786},
      archivePrefix={arXiv},
      primaryClass={cs.CV},
      url={https://arxiv.org/abs/2305.13786}, 
}

@misc{shao2024deepseekmathpushinglimitsmathematical,
      title={DeepSeekMath: Pushing the Limits of Mathematical Reasoning in Open Language Models}, 
      author={Zhihong Shao and Peiyi Wang and Qihao Zhu and Runxin Xu and Junxiao Song and Xiao Bi and Haowei Zhang and Mingchuan Zhang and Y. K. Li and Y. Wu and Daya Guo},
      year={2024},
      eprint={2402.03300},
      archivePrefix={arXiv},
      primaryClass={cs.CL},
      url={https://arxiv.org/abs/2402.03300}, 
}

@misc{lightman2023letsverifystepstep,
      title={Let's Verify Step by Step}, 
      author={Hunter Lightman and Vineet Kosaraju and Yura Burda and Harri Edwards and Bowen Baker and Teddy Lee and Jan Leike and John Schulman and Ilya Sutskever and Karl Cobbe},
      year={2023},
      eprint={2305.20050},
      archivePrefix={arXiv},
      primaryClass={cs.LG},
      url={https://arxiv.org/abs/2305.20050}, 
}

@misc{luo2025wizardmathempoweringmathematicalreasoning,
      title={WizardMath: Empowering Mathematical Reasoning for Large Language Models via Reinforced Evol-Instruct}, 
      author={Haipeng Luo and Qingfeng Sun and Can Xu and Pu Zhao and Jianguang Lou and Chongyang Tao and Xiubo Geng and Qingwei Lin and Shifeng Chen and Yansong Tang and Dongmei Zhang},
      year={2025},
      eprint={2308.09583},
      archivePrefix={arXiv},
      primaryClass={cs.CL},
      url={https://arxiv.org/abs/2308.09583}, 
}

@misc{deepseekai2025deepseekr1incentivizingreasoningcapability,
      title={DeepSeek-R1: Incentivizing Reasoning Capability in LLMs via Reinforcement Learning}, 
      author={DeepSeek-AI and Daya Guo and Dejian Yang and Haowei Zhang and Junxiao Song and Ruoyu Zhang and Runxin Xu and Qihao Zhu and Shirong Ma and Peiyi Wang and Xiao Bi and Xiaokang Zhang and Xingkai Yu and Yu Wu and Z. F. Wu and Zhibin Gou and Zhihong Shao and Zhuoshu Li and Ziyi Gao and Aixin Liu and Bing Xue and Bingxuan Wang and Bochao Wu and Bei Feng and Chengda Lu and Chenggang Zhao and Chengqi Deng and Chenyu Zhang and Chong Ruan and Damai Dai and Deli Chen and Dongjie Ji and Erhang Li and Fangyun Lin and Fucong Dai and Fuli Luo and Guangbo Hao and Guanting Chen and Guowei Li and H. Zhang and Han Bao and Hanwei Xu and Haocheng Wang and Honghui Ding and Huajian Xin and Huazuo Gao and Hui Qu and Hui Li and Jianzhong Guo and Jiashi Li and Jiawei Wang and Jingchang Chen and Jingyang Yuan and Junjie Qiu and Junlong Li and J. L. Cai and Jiaqi Ni and Jian Liang and Jin Chen and Kai Dong and Kai Hu and Kaige Gao and Kang Guan and Kexin Huang and Kuai Yu and Lean Wang and Lecong Zhang and Liang Zhao and Litong Wang and Liyue Zhang and Lei Xu and Leyi Xia and Mingchuan Zhang and Minghua Zhang and Minghui Tang and Meng Li and Miaojun Wang and Mingming Li and Ning Tian and Panpan Huang and Peng Zhang and Qiancheng Wang and Qinyu Chen and Qiushi Du and Ruiqi Ge and Ruisong Zhang and Ruizhe Pan and Runji Wang and R. J. Chen and R. L. Jin and Ruyi Chen and Shanghao Lu and Shangyan Zhou and Shanhuang Chen and Shengfeng Ye and Shiyu Wang and Shuiping Yu and Shunfeng Zhou and Shuting Pan and S. S. Li and Shuang Zhou and Shaoqing Wu and Shengfeng Ye and Tao Yun and Tian Pei and Tianyu Sun and T. Wang and Wangding Zeng and Wanjia Zhao and Wen Liu and Wenfeng Liang and Wenjun Gao and Wenqin Yu and Wentao Zhang and W. L. Xiao and Wei An and Xiaodong Liu and Xiaohan Wang and Xiaokang Chen and Xiaotao Nie and Xin Cheng and Xin Liu and Xin Xie and Xingchao Liu and Xinyu Yang and Xinyuan Li and Xuecheng Su and Xuheng Lin and X. Q. Li and Xiangyue Jin and Xiaojin Shen and Xiaosha Chen and Xiaowen Sun and Xiaoxiang Wang and Xinnan Song and Xinyi Zhou and Xianzu Wang and Xinxia Shan and Y. K. Li and Y. Q. Wang and Y. X. Wei and Yang Zhang and Yanhong Xu and Yao Li and Yao Zhao and Yaofeng Sun and Yaohui Wang and Yi Yu and Yichao Zhang and Yifan Shi and Yiliang Xiong and Ying He and Yishi Piao and Yisong Wang and Yixuan Tan and Yiyang Ma and Yiyuan Liu and Yongqiang Guo and Yuan Ou and Yuduan Wang and Yue Gong and Yuheng Zou and Yujia He and Yunfan Xiong and Yuxiang Luo and Yuxiang You and Yuxuan Liu and Yuyang Zhou and Y. X. Zhu and Yanhong Xu and Yanping Huang and Yaohui Li and Yi Zheng and Yuchen Zhu and Yunxian Ma and Ying Tang and Yukun Zha and Yuting Yan and Z. Z. Ren and Zehui Ren and Zhangli Sha and Zhe Fu and Zhean Xu and Zhenda Xie and Zhengyan Zhang and Zhewen Hao and Zhicheng Ma and Zhigang Yan and Zhiyu Wu and Zihui Gu and Zijia Zhu and Zijun Liu and Zilin Li and Ziwei Xie and Ziyang Song and Zizheng Pan and Zhen Huang and Zhipeng Xu and Zhongyu Zhang and Zhen Zhang},
      year={2025},
      eprint={2501.12948},
      archivePrefix={arXiv},
      primaryClass={cs.CL},
      url={https://arxiv.org/abs/2501.12948}, 
}

@misc{lee2024vhelmholisticevaluationvision,
      title={VHELM: A Holistic Evaluation of Vision Language Models}, 
      author={Tony Lee and Haoqin Tu and Chi Heem Wong and Wenhao Zheng and Yiyang Zhou and Yifan Mai and Josselin Somerville Roberts and Michihiro Yasunaga and Huaxiu Yao and Cihang Xie and Percy Liang},
      year={2024},
      eprint={2410.07112},
      archivePrefix={arXiv},
      primaryClass={cs.CV},
      url={https://arxiv.org/abs/2410.07112}, 
}

@misc{bai2025qwen25vltechnicalreport,
      title={Qwen2.5-VL Technical Report}, 
      author={Shuai Bai and Keqin Chen and Xuejing Liu and Jialin Wang and Wenbin Ge and Sibo Song and Kai Dang and Peng Wang and Shijie Wang and Jun Tang and Humen Zhong and Yuanzhi Zhu and Mingkun Yang and Zhaohai Li and Jianqiang Wan and Pengfei Wang and Wei Ding and Zheren Fu and Yiheng Xu and Jiabo Ye and Xi Zhang and Tianbao Xie and Zesen Cheng and Hang Zhang and Zhibo Yang and Haiyang Xu and Junyang Lin},
      year={2025},
      eprint={2502.13923},
      archivePrefix={arXiv},
      primaryClass={cs.CV},
      url={https://arxiv.org/abs/2502.13923}, 
}

@misc{qwen3technicalreport,
      title={Qwen3 Technical Report}, 
      author={An Yang and Anfeng Li and Baosong Yang and Beichen Zhang and Binyuan Hui and Bo Zheng and Bowen Yu and Chang Gao and Chengen Huang and Chenxu Lv and Chujie Zheng and Dayiheng Liu and Fan Zhou and Fei Huang and Feng Hu and Hao Ge and Haoran Wei and Huan Lin and Jialong Tang and Jian Yang and Jianhong Tu and Jianwei Zhang and Jianxin Yang and Jiaxi Yang and Jing Zhou and Jingren Zhou and Junyang Lin and Kai Dang and Keqin Bao and Kexin Yang and Le Yu and Lianghao Deng and Mei Li and Mingfeng Xue and Mingze Li and Pei Zhang and Peng Wang and Qin Zhu and Rui Men and Ruize Gao and Shixuan Liu and Shuang Luo and Tianhao Li and Tianyi Tang and Wenbiao Yin and Xingzhang Ren and Xinyu Wang and Xinyu Zhang and Xuancheng Ren and Yang Fan and Yang Su and Yichang Zhang and Yinger Zhang and Yu Wan and Yuqiong Liu and Zekun Wang and Zeyu Cui and Zhenru Zhang and Zhipeng Zhou and Zihan Qiu},
      year={2025},
      eprint={2505.09388},
      archivePrefix={arXiv},
      primaryClass={cs.CL},
      url={https://arxiv.org/abs/2505.09388}, 
}

@inproceedings{luo2024serl,
  title={Serl: A software suite for sample-efficient robotic reinforcement learning},
  author={Luo, Jianlan and Hu, Zheyuan and Xu, Charles and Tan, You Liang and Berg, Jacob and Sharma, Archit and Schaal, Stefan and Finn, Chelsea and Gupta, Abhishek and Levine, Sergey},
  booktitle={2024 IEEE International Conference on Robotics and Automation (ICRA)},
  pages={16961--16969},
  year={2024},
  organization={IEEE}
}

@article{luo2025precise,
  title={Precise and dexterous robotic manipulation via human-in-the-loop reinforcement learning},
  author={Luo, Jianlan and Xu, Charles and Wu, Jeffrey and Levine, Sergey},
  journal={Science Robotics},
  volume={10},
  number={105},
  pages={eads5033},
  year={2025},
  publisher={American Association for the Advancement of Science}
}

@article{smith2022walk,
  title={A walk in the park: Learning to walk in 20 minutes with model-free reinforcement learning},
  author={Smith, Laura and Kostrikov, Ilya and Levine, Sergey},
  journal={arXiv preprint arXiv:2208.07860},
  year={2022}
}

@article{mark2024policy,
  title={Policy agnostic rl: Offline rl and online rl fine-tuning of any class and backbone},
  author={Mark, Max Sobol and Gao, Tian and Sampaio, Georgia Gabriela and Srirama, Mohan Kumar and Sharma, Archit and Finn, Chelsea and Kumar, Aviral},
  journal={arXiv preprint arXiv:2412.06685},
  year={2024}
}

@inproceedings{ankile2025imitation,
  title={From imitation to refinement-residual rl for precise assembly},
  author={Ankile, Lars and Simeonov, Anthony and Shenfeld, Idan and Torne, Marcel and Agrawal, Pulkit},
  booktitle={2025 IEEE International Conference on Robotics and Automation (ICRA)},
  pages={01--08},
  year={2025},
  organization={IEEE}
}

@article{chen2025conrft,
  title={Conrft: A reinforced fine-tuning method for vla models via consistency policy},
  author={Chen, Yuhui and Tian, Shuai and Liu, Shugao and Zhou, Yingting and Li, Haoran and Zhao, Dongbin},
  journal={arXiv preprint arXiv:2502.05450},
  year={2025}
}

@article{openxembodiment2023,
  title   = {Open X-Embodiment: Robotic Learning Datasets and RT-X Models},
  author  = {{Open X-Embodiment Collaboration}},
  journal = {arXiv preprint arXiv:2310.08864},
  year    = {2023},
  doi     = {10.48550/arXiv.2310.08864},
  url     = {https://arxiv.org/abs/2310.08864}
}

@article{atreya2025roboarena,
  title   = {RoboArena: Distributed Real-World Evaluation of Generalist Robot Policies},
  author  = {Pranav Atreya and Karl Pertsch and Tony Lee and Moo Jin Kim and Arhan Jain and Artur Kuramshin and Clemens Eppner and Cyrus Neary and Edward Hu and Fabio Ramos and Jonathan Tremblay and Kanav Arora and Kirsty Ellis and Luca Macesanu and Matthew Leonard and Meedeum Cho and Ozgur Aslan and Shivin Dass and Jie Wang and Xingfang Yuan and Xuning Yang and Abhishek Gupta and Dinesh Jayaraman and Glen Berseth and Kostas Daniilidis and Roberto Mart{\'\i}n-Mart{\'\i}n and Youngwoon Lee and Percy Liang and Chelsea Finn and Sergey Levine},
  journal = {arXiv preprint arXiv:2506.18123},
  year    = {2025},
  url     = {https://arxiv.org/abs/2506.18123}
}

@article{ma2024gvl,
  title   = {Vision Language Models are In-Context Value Learners},
  author  = {Ma, Yecheng Jason and Hejna, Joey and Wahid, Ayzaan and Fu, Chuyuan and Shah, Dhruv and Liang, Jacky and Xu, Zhuo and Kirmani, Sean and Xu, Peng and Driess, Danny and Xiao, Ted and Tompson, Jonathan and Bastani, Osbert and Jayaraman, Dinesh and Yu, Wenhao and Zhang, Tingnan and Sadigh, Dorsa and Xia, Fei},
  journal = {arXiv preprint arXiv:2411.04549},
  year    = {2024},
  url     = {https://arxiv.org/abs/2411.04549}
}

@misc{opengvl2025,
  title        = {OpenGVL: Task Completion Leaderboard for Evaluating VLMs as Temporal Value Estimators},
  author       = {{OpenGVL Team}},
  howpublished = {\url{https://huggingface.co/spaces/OpenGVL/OpenGVL}},
  note         = {Accessed: 2025-09-04},
  year         = {2025}
}

@article{lambert2024rewardbench,
  title        = {RewardBench: Evaluating Reward Models for Language Modeling},
  author       = {Nathan Lambert and Valentina Pyatkin and Jacob Morrison and
                  L.~J. Miranda and Bill Yuchen Lin and Khyathi Chandu and
                  Nouha Dziri and Sachin Kumar and Tom Zick and Yejin Choi and
                  Noah A. Smith and Hannaneh Hajishirzi},
  journal      = {arXiv preprint arXiv:2403.13787},
  year         = {2024},
  url          = {https://arxiv.org/abs/2403.13787}
}

@article{malik2025rewardbench2,
  title        = {RewardBench 2: Advancing Reward Model Evaluation},
  author       = {Saumya Malik and Valentina Pyatkin and Sander Land and
                  Jacob Morrison and Noah A. Smith and Hannaneh Hajishirzi and
                  Nathan Lambert},
  journal      = {arXiv preprint arXiv:2506.01937},
  year         = {2025},
  url          = {https://arxiv.org/abs/2506.01937}
}

@article{li2024vlrewardbench,
  title        = {VLRewardBench: A Challenging Benchmark for Vision--Language
                  Generative Reward Models},
  author       = {Lei Li and Yuancheng Wei and Zhihui Xie and Xuqing Yang and
                  Yifan Song and Peiyi Wang and Chenxin An and Tianyu Liu and
                  Sujian Li and Bill Yuchen Lin and Lingpeng Kong and Qi Liu},
  journal      = {arXiv preprint arXiv:2411.17451},
  year         = {2024},
  url          = {https://arxiv.org/abs/2411.17451}
}

@article{yasunaga2025mmrewardbench,
  title        = {Multimodal RewardBench: Holistic Evaluation of Reward Models
                  for Vision--Language Models},
  author       = {Michihiro Yasunaga and Luke Zettlemoyer and
                  Marjan Ghazvininejad},
  journal      = {arXiv preprint arXiv:2502.14191},
  year         = {2025},
  url          = {https://arxiv.org/abs/2502.14191}
}

@article{yang2024adapt2reward,
  title   = {Adapt2Reward: Adapting Video–Language Models to Generalizable Robotic Rewards via Failure Prompts},
  author  = {Yang, Yanting and Chen, Minghao and Qiu, Qibo and Wu, Jiahao and Wang, Wenxiao and Lin, Binbin and Guan, Ziyu and He, Xiaofei},
  journal = {arXiv preprint arXiv:2407.14872},
  year    = {2024},
  url     = {https://arxiv.org/abs/2407.14872}
}

@article{alakuijala2024videolanguagecritic,
  title   = {Video-Language Critic: Transferable Reward Functions for Language-Conditioned Robotics},
  author  = {Alakuijala, Minttu and McLean, Reginald and Woungang, Isaac and Farsad, Nariman and Kaski, Samuel and Marttinen, Pekka and Yuan, Kai},
  journal = {arXiv preprint arXiv:2405.19988},
  year    = {2024},
  url     = {https://arxiv.org/abs/2405.19988}
}

@article{patel2025realtosim,
  title   = {A Real-to-Sim-to-Real Approach to Robotic Manipulation with VLM-Generated Iterative Keypoint Rewards},
  author  = {Patel, Shivansh and Yin, Xinchen and Huang, Wenlong and Garg, Shubham and Nayyeri, Hooshang and Fei-Fei, Li and Lazebnik, Svetlana and Li, Yunzhu},
  journal = {arXiv preprint arXiv:2502.08643},
  year    = {2025},
  url     = {https://arxiv.org/abs/2502.08643}
}

@article{zeng2024video2reward,
  title   = {Video2Reward: Generating Reward Function from Videos for Legged Robot Behavior Learning},
  author  = {Zeng, Runhao and Zhou, Dingjie and Liang, Qiwei and Liu, Junlin and Li, Hui and Huang, Changxin and Li, Jianqiang and Hu, Xiping and Sun, Fuchun},
  journal = {arXiv preprint arXiv:2412.05515},
  year    = {2024},
  url     = {https://arxiv.org/abs/2412.05515}
}

@article{yang2024rank2reward,
  title   = {Rank2Reward: Learning Shaped Reward Functions from Passive Video},
  author  = {Yang, Daniel and Tjia, Davin and Berg, Jacob and Damen, Dima and Agrawal, Pulkit and Gupta, Abhishek},
  journal = {arXiv preprint arXiv:2404.14735},
  year    = {2024},
  url     = {https://arxiv.org/abs/2404.14735}
}

@article{wang2024rlvlmf,
  title   = {RL-VLM-F: Reinforcement Learning from Vision–Language Foundation Model Feedback},
  author  = {Yufei Wang and Zhanyi Sun and Jesse Zhang and Zhou Xian and Erdem Biyik and David Held and Zackory Erickson},
  journal = {arXiv preprint arXiv:2402.03681},
  year    = {2024},
  url     = {https://arxiv.org/abs/2402.03681}
}

@article{venkataraman2024offlinevlm,
  title   = {Real-World Offline Reinforcement Learning from Vision Language Model Feedback},
  author  = {Sreyas Venkataraman and Yufei Wang and Ziyu Wang and Zackory Erickson and David Held},
  journal = {arXiv preprint arXiv:2411.05273},
  year    = {2024},
  url     = {https://arxiv.org/abs/2411.05273}
}

@article{chen2025tevir,
  title   = {TeViR: Text-to-Video Reward with Diffusion Models for Efficient Reinforcement Learning},
  author  = {Yuhui Chen and Haoran Li and Zhennan Jiang and Haowei Wen and Dongbin Zhao},
  journal = {arXiv preprint arXiv:2505.19769},
  year    = {2025},
  url     = {https://arxiv.org/abs/2505.19769}
}

@article{luu2025erlvlm,
  title   = {ERL-VLM: Enhancing Rating-Based Reinforcement Learning to Effectively Leverage Feedback from Large Vision–Language Models},
  author  = {Tung M. Luu and Younghwan Lee and Donghoon Lee and Sunho Kim and Min Jun Kim and Chang D. Yoo},
  journal = {arXiv preprint arXiv:2506.12822},
  year    = {2025},
  url     = {https://arxiv.org/abs/2506.12822}
}

@article{singh2025varp,
  title   = {VARP: Reinforcement Learning from Vision–Language Model Feedback with Agent-Regularized Preferences},
  author  = {Anukriti Singh and Amisha Bhaskar and Peihong Yu and Souradip Chakraborty and Ruthwik Dasyam and Amrit Bedi and Pratap Tokekar},
  journal = {arXiv preprint arXiv:2503.13817},
  year    = {2025},
  url     = {https://arxiv.org/pdf/2503.13817}
}

@article{huang2024vlmrl,
  title   = {VLM-RL: A Unified Vision–Language Model and Reinforcement Learning Framework for Safe Autonomous Driving},
  author  = {Zilin Huang and Zihao Sheng and Yansong Qu and Junwei You and Sikai Chen},
  journal = {arXiv preprint arXiv:2412.15544},
  year    = {2024},
  url     = {https://arxiv.org/abs/2412.15544}
}

@article{zhang2025rewind,
  title   = {ReWiND: Language-Guided Rewards Teach Robot Policies without New Demonstrations},
  author  = {Jiahui Zhang and Yusen Luo and Abrar Anwar and Sumedh A. Sontakke and Joseph J. Lim and Jesse Thomason and Erdem Bıyık and Jesse Zhang},
  journal = {arXiv preprint arXiv:2505.10911},
  year    = {2025},
  url     = {https://arxiv.org/abs/2505.10911}
}

@inproceedings{walke2023bridgedata,
  title={BridgeData V2: A Dataset for Robot Learning at Scale},
  author={Walke, Homer and Black, Kevin and Lee, Abraham and Kim, Moo Jin and Du, Max and Zheng, Chongyi and Zhao, Tony and Hansen-Estruch, Philippe and Vuong, Quan and He, Andre and Myers, Vivek and Fang, Kuan and Finn, Chelsea and Levine, Sergey},
  booktitle={Conference on Robot Learning (CoRL)},
  year={2023}
}

@inproceedings{rosete2022tacorl,
  author = {Rosete-Beas, Erick and Mees, Oier and Kalweit, Gabriel and Boedecker, Joschka and Burgard, Wolfram},
  title = {Latent Plans for Task Agnostic Offline Reinforcement Learning},
  booktitle = {Conference on Robot Learning (CoRL)},
  year = {2022}
}

@inproceedings{mees23hulc2,
  title={Grounding Language with Visual Affordances over Unstructured Data},
  author={Mees, Oier and Borja-Diaz, Jessica and Burgard, Wolfram},
  booktitle={IEEE International Conference on Robotics and Automation (ICRA)},
  year={2023}
}

@software{dass2023jacoplay,
  author = {Dass, Shivin and Yapeter, Jullian and Zhang, Jesse and Zhang, Jiahui and Pertsch, Karl and Nikolaidis, Stefanos and Lim, Joseph J.},
  title = {CLVR Jaco Play Dataset},
  url = {https://github.com/clvrai/clvr_jaco_play_dataset},
  year = {2023}
}

@inproceedings{mandlekar2019scaling,
  title={Scaling Robot Supervision to Hundreds of Hours with RoboTurk: Robotic Manipulation Dataset Through Human Reasoning and Dexterity},
  author={Mandlekar, Ajay and Booher, Jonathan and Spero, Max and Tung, Albert and Gupta, Anchit and Zhu, Yuke and Garg, Animesh and Savarese, Silvio and Fei-Fei, Li},
  booktitle={IEEE/RSJ International Conference on Intelligent Robots and Systems (IROS)},
  pages={1048--1055},
  year={2019}
}

@article{pari2021surprising,
  title={The Surprising Effectiveness of Representation Learning for Visual Imitation},
  author={Pari, Jyothish and Shafiullah, Nur Muhammad and Arunachalam, Sridhar Pandian and Pinto, Lerrel},
  journal={arXiv preprint arXiv:2112.01511},
  year={2021}
}

@inproceedings{zhu2022viola,
  title={VIOLA: Imitation Learning for Vision-Based Manipulation with Object Proposal Priors},
  author={Zhu, Yifeng and Joshi, Abhishek and Stone, Peter and Zhu, Yuke},
  booktitle={Conference on Robot Learning (CoRL)},
  year={2022}
}

@misc{BerkeleyUR5Website,
  title = {Berkeley UR5 Demonstration Dataset},
  author = {Chen, Lawrence Yunliang and Adebola, Simeon and Goldberg, Ken},
  howpublished = {\url{https://sites.google.com/view/berkeley-ur5/home}}
}

@inproceedings{belkhale2023hydra,
  title={HYDRA: Hybrid Robot Actions for Imitation Learning},
  author={Belkhale, Suneel and Cui, Yuchen and Sadigh, Dorsa},
  booktitle={Robotics: Science and Systems (RSS)},
  year={2023}
}

@article{zhu2022bottom,
  title={Bottom-Up Skill Discovery From Unsegmented Demonstrations for Long-Horizon Robot Manipulation},
  author={Zhu, Yifeng and Stone, Peter and Zhu, Yuke},
  journal={IEEE Robotics and Automation Letters},
  volume={7},
  number={2},
  pages={4126--4133},
  year={2022}
}

@misc{ucsd_kitchens,
  author = {Yan, Ge and Wu, Kris and Wang, Xiaolong},
  title = {UCSD Kitchens Dataset},
  year = {2023},
  howpublished = {\url{https://vis-www.cs.umich.edu/ucsd-kitchens}}
}

@misc{oh2023pr2utokyodatasets,
  author = {Oh, Jihoon and Kanazawa, Naoaki and Kawaharazuka, Kento},
  title = {X-Embodiment U-Tokyo PR2 Datasets},
  year = {2023},
  howpublished = {\url{https://github.com/ojh6404/rlds_dataset_builder}}
}

@misc{matsushima2023weblab,
  title        = {WebLab xArm Datasets},
  author       = {Matsushima, Tatsuya and Furuta, Hiroki and Iwasawa, Yusuke and Matsuo, Yutaka},
  year         = {2023},
  howpublished = {\url{https://github.com/weblab-xarm}}
}

@article{salhotra2022dmfd,
  title={Learning Deformable Object Manipulation From Expert Demonstrations},
  author={Salhotra, Gautam and Liu, I-Chun Arthur and Dominguez-Kuhne, Marcus and Sukhatme, Gaurav S.},
  journal={IEEE Robotics and Automation Letters},
  volume={7},
  number={4},
  pages={8775--8782},
  year={2022}
}

@inproceedings{Radosavovic2022,
  title = {Real-World Robot Learning with Masked Visual Pre-Training},
  author = {Radosavovic, Ilija and Xiao, Tete and James, Stephen and Abbeel, Pieter and Malik, Jitendra and Darrell, Trevor},
  booktitle = {Conference on Robot Learning (CoRL)},
  year = {2022}
}

@article{Radosavovic2023,
  title={Robot Learning with Sensorimotor Pre-Training},
  author={Radosavovic, Ilija and Shi, Baifeng and Fu, Letian and Goldberg, Ken and Darrell, Trevor and Malik, Jitendra},
  journal={arXiv preprint arXiv:2306.10007},
  year={2023}
}

@inproceedings{haldar2023watch,
  title={Watch and Match: Supercharging Imitation with Regularized Optimal Transport},
  author={Haldar, Siddhant and Mathur, Vaibhav and Yarats, Denis and Pinto, Lerrel},
  booktitle={Conference on Robot Learning (PMLR)},
  year={2023}
}

@article{Feng2023Finetuning,
  title={Finetuning Offline World Models in the Real World},
  author={Feng, Yunhai and Hansen, Nicklas and Xiong, Ziyan and Rajagopalan, Chandramouli and Wang, Xiaolong},
  journal={arXiv preprint arXiv:2312.00000},
  year={2023}
}

@inproceedings{dasari2025ingredients,
  title={The ingredients for robotic diffusion transformers},
  author={Dasari, Sudeep and Mees, Oier and Zhao, Sebastian and Srirama, Mohan Kumar and Levine, Sergey},
  booktitle={2025 IEEE International Conference on Robotics and Automation (ICRA)},
  pages={15617--15625},
  year={2025},
  organization={IEEE}
}

@article{nakamoto2024steering,
  title={Steering your generalists: Improving robotic foundation models via value guidance},
  author={Nakamoto, Mitsuhiko and Mees, Oier and Kumar, Aviral and Levine, Sergey},
  journal={arXiv preprint arXiv:2410.13816},
  year={2024}
}

@article{zhang2024grape,
  title={Grape: Generalizing robot policy via preference alignment},
  author={Zhang, Zijian and Zheng, Kaiyuan and Chen, Zhaorun and Jang, Joel and Li, Yi and Han, Siwei and Wang, Chaoqi and Ding, Mingyu and Fox, Dieter and Yao, Huaxiu},
  journal={arXiv preprint arXiv:2411.19309},
  year={2024}
}

@article{dong2025matters,
  title={What Matters for Batch Online Reinforcement Learning in Robotics?},
  author={Dong, Perry and Mirchandani, Suvir and Sadigh, Dorsa and Finn, Chelsea},
  journal={arXiv preprint arXiv:2505.08078},
  year={2025}
}

@inproceedings{hu2025flare,
  title={Flare: Achieving masterful and adaptive robot policies with large-scale reinforcement learning fine-tuning},
  author={Hu, Jiaheng and Hendrix, Rose and Farhadi, Ali and Kembhavi, Aniruddha and Mart{\'\i}n-Mart{\'\i}n, Roberto and Stone, Peter and Zeng, Kuo-Hao and Ehsani, Kiana},
  booktitle={2025 IEEE International Conference on Robotics and Automation (ICRA)},
  pages={3617--3624},
  year={2025},
  organization={IEEE}
}

@article{kumar2021rma,
  title={Rma: Rapid motor adaptation for legged robots},
  author={Kumar, Ashish and Fu, Zipeng and Pathak, Deepak and Malik, Jitendra},
  journal={arXiv preprint arXiv:2107.04034},
  year={2021}
}

@article{zhu2020ingredients,
  title={The ingredients of real-world robotic reinforcement learning},
  author={Zhu, Henry and Yu, Justin and Gupta, Abhishek and Shah, Dhruv and Hartikainen, Kristian and Singh, Avi and Kumar, Vikash and Levine, Sergey},
  journal={arXiv preprint arXiv:2004.12570},
  year={2020}
}

@article{mendonca2024continuously,
  title={Continuously improving mobile manipulation with autonomous real-world rl},
  author={Mendonca, Russell and Panov, Emmanuel and Bucher, Bernadette and Wang, Jiuguang and Pathak, Deepak},
  journal={arXiv preprint arXiv:2409.20568},
  year={2024}
}

@article{riedmiller2009reinforcement,
  title={Reinforcement learning for robot soccer},
  author={Riedmiller, Martin and Gabel, Thomas and Hafner, Roland and Lange, Sascha},
  journal={Autonomous Robots},
  volume={27},
  number={1},
  pages={55--73},
  year={2009},
  publisher={Springer}
}

@inproceedings{cutler2014reinforcement,
  title={Reinforcement learning with multi-fidelity simulators},
  author={Cutler, Mark and Walsh, Thomas J and How, Jonathan P},
  booktitle={2014 IEEE International Conference on Robotics and Automation (ICRA)},
  pages={3888--3895},
  year={2014},
  organization={IEEE}
}

@inproceedings{tobin2017domain,
  title={Domain randomization for transferring deep neural networks from simulation to the real world},
  author={Tobin, Josh and Fong, Rachel and Ray, Alex and Schneider, Jonas and Zaremba, Wojciech and Abbeel, Pieter},
  booktitle={2017 IEEE/RSJ international conference on intelligent robots and systems (IROS)},
  pages={23--30},
  year={2017},
  organization={IEEE}
}

@inproceedings{peng2018sim,
  title={Sim-to-real transfer of robotic control with dynamics randomization},
  author={Peng, Xue Bin and Andrychowicz, Marcin and Zaremba, Wojciech and Abbeel, Pieter},
  booktitle={2018 IEEE international conference on robotics and automation (ICRA)},
  pages={3803--3810},
  year={2018},
  organization={IEEE}
}

@article{rajeswaran2016epopt,
  title={Epopt: Learning robust neural network policies using model ensembles},
  author={Rajeswaran, Aravind and Ghotra, Sarvjeet and Ravindran, Balaraman and Levine, Sergey},
  journal={arXiv preprint arXiv:1610.01283},
  year={2016}
}

@inproceedings{chebotar2019closing,
    title={Closing the sim-to-real loop: Adapting simulation randomization with real world experience},
    author={Chebotar, Yevgen and Handa, Ankur and Makoviychuk, Viktor and Macklin, Miles and Issac, Jan and Ratliff, Nathan and Fox, Dieter},
    booktitle={ICRA},
    year={2019},
}

@inproceedings{smith2022legged,
  title={Legged robots that keep on learning: Fine-tuning locomotion policies in the real world},
  author={Smith, Laura and Kew, J Chase and Peng, Xue Bin and Ha, Sehoon and Tan, Jie and Levine, Sergey},
  booktitle={2022 international conference on robotics and automation (ICRA)},
  pages={1593--1599},
  year={2022},
  organization={IEEE}
}

@article{black2024pi_0,
  title={$pi\_0 $: A Vision-Language-Action Flow Model for General Robot Control},
  author={Black, Kevin and Brown, Noah and Driess, Danny and Esmail, Adnan and Equi, Michael and Finn, Chelsea and Fusai, Niccolo and Groom, Lachy and Hausman, Karol and Ichter, Brian and others},
  journal={arXiv preprint arXiv:2410.24164},
  year={2024}
}

@article{kim2024openvla,
  title={OpenVLA: An Open-Source Vision-Language-Action Model},
  author={Kim, Moo Jin and Pertsch, Karl and Karamcheti, Siddharth and Xiao, Ted and Balakrishna, Ashwin and Nair, Suraj and Rafailov, Rafael and Foster, Ethan and Lam, Grace and Sanketi, Pannag and others},
  journal={arXiv preprint arXiv:2406.09246},
  year={2024}
}

@inproceedings{octo_2023,
    title={Octo: An Open-Source Generalist Robot Policy},
    author = {{Octo Model Team} and Dibya Ghosh and Homer Walke and Karl Pertsch and Kevin Black and Oier Mees and Sudeep Dasari and Joey Hejna and Charles Xu and Jianlan Luo and Tobias Kreiman and {You Liang} Tan and Pannag Sanketi and Quan Vuong and Ted Xiao and Dorsa Sadigh and Chelsea Finn and Sergey Levine},
    booktitle = {Proceedings of Robotics: Science and Systems},
    address  = {Delft, Netherlands},
    year = {2024},
}

@misc{open_x_embodiment_rt_x_2023,
title={Open {X-E}mbodiment: Robotic Learning Datasets and {RT-X} Models},
author = {{Open X-Embodiment Collaboration} and Abhishek Padalkar and Acorn Pooley and Ajinkya Jain and Alex Bewley and Alex Herzog and Alex Irpan and Alexander Khazatsky and Anant Rai and Anikait Singh and Anthony Brohan and Antonin Raffin and Ayzaan Wahid and Ben Burgess-Limerick and Beomjoon Kim and Bernhard Schölkopf and Brian Ichter and Cewu Lu and Charles Xu and Chelsea Finn and Chenfeng Xu and Cheng Chi and Chenguang Huang and Christine Chan and Chuer Pan and Chuyuan Fu and Coline Devin and Danny Driess and Deepak Pathak and Dhruv Shah and Dieter Büchler and Dmitry Kalashnikov and Dorsa Sadigh and Edward Johns and Federico Ceola and Fei Xia and Freek Stulp and Gaoyue Zhou and Gaurav S. Sukhatme and Gautam Salhotra and Ge Yan and Giulio Schiavi and Hao Su and Hao-Shu Fang and Haochen Shi and Heni Ben Amor and Henrik I Christensen and Hiroki Furuta and Homer Walke and Hongjie Fang and Igor Mordatch and Ilija Radosavovic and Isabel Leal and Jacky Liang and Jaehyung Kim and Jan Schneider and Jasmine Hsu and Jeannette Bohg and Jeffrey Bingham and Jiajun Wu and Jialin Wu and Jianlan Luo and Jiayuan Gu and Jie Tan and Jihoon Oh and Jitendra Malik and Jonathan Tompson and Jonathan Yang and Joseph J. Lim and João Silvério and Junhyek Han and Kanishka Rao and Karl Pertsch and Karol Hausman and Keegan Go and Keerthana Gopalakrishnan and Ken Goldberg and Kendra Byrne and Kenneth Oslund and Kento Kawaharazuka and Kevin Zhang and Keyvan Majd and Krishan Rana and Krishnan Srinivasan and Lawrence Yunliang Chen and Lerrel Pinto and Liam Tan and Lionel Ott and Lisa Lee and Masayoshi Tomizuka and Maximilian Du and Michael Ahn and Mingtong Zhang and Mingyu Ding and Mohan Kumar Srirama and Mohit Sharma and Moo Jin Kim and Naoaki Kanazawa and Nicklas Hansen and Nicolas Heess and Nikhil J Joshi and Niko Suenderhauf and Norman Di Palo and Nur Muhammad Mahi Shafiullah and Oier Mees and Oliver Kroemer and Pannag R Sanketi and Paul Wohlhart and Peng Xu and Pierre Sermanet and Priya Sundaresan and Quan Vuong and Rafael Rafailov and Ran Tian and Ria Doshi and Roberto Martín-Martín and Russell Mendonca and Rutav Shah and Ryan Hoque and Ryan Julian and Samuel Bustamante and Sean Kirmani and Sergey Levine and Sherry Moore and Shikhar Bahl and Shivin Dass and Shuran Song and Sichun Xu and Siddhant Haldar and Simeon Adebola and Simon Guist and Soroush Nasiriany and Stefan Schaal and Stefan Welker and Stephen Tian and Sudeep Dasari and Suneel Belkhale and Takayuki Osa and Tatsuya Harada and Tatsuya Matsushima and Ted Xiao and Tianhe Yu and Tianli Ding and Todor Davchev and Tony Z. Zhao and Travis Armstrong and Trevor Darrell and Vidhi Jain and Vincent Vanhoucke and Wei Zhan and Wenxuan Zhou and Wolfram Burgard and Xi Chen and Xiaolong Wang and Xinghao Zhu and Xuanlin Li and Yao Lu and Yevgen Chebotar and Yifan Zhou and Yifeng Zhu and Ying Xu and Yixuan Wang and Yonatan Bisk and Yoonyoung Cho and Youngwoon Lee and Yuchen Cui and Yueh-hua Wu and Yujin Tang and Yuke Zhu and Yunzhu Li and Yusuke Iwasawa and Yutaka Matsuo and Zhuo Xu and Zichen Jeff Cui},
howpublished  = {\url{https://arxiv.org/abs/2310.08864}},
year = {2023},
}

@inproceedings{khazatsky2024droid,
    title   = {DROID: A Large-Scale In-The-Wild Robot Manipulation Dataset},
    author  = {Alexander Khazatsky and Karl Pertsch and Suraj Nair and Ashwin Balakrishna and Sudeep Dasari and Siddharth Karamcheti and Soroush Nasiriany and Mohan Kumar Srirama and Lawrence Yunliang Chen and Kirsty Ellis and Peter David Fagan and Joey Hejna and Masha Itkina and Marion Lepert and Yecheng Jason Ma and Patrick Tree Miller and Jimmy Wu and Suneel Belkhale and Shivin Dass and Huy Ha and Arhan Jain and Abraham Lee and Youngwoon Lee and Marius Memmel and Sungjae Park and Ilija Radosavovic and Kaiyuan Wang and Albert Zhan and Kevin Black and Cheng Chi and Kyle Beltran Hatch and Shan Lin and Jingpei Lu and Jean Mercat and Abdul Rehman and Pannag R Sanketi and Archit Sharma and Cody Simpson and Quan Vuong and Homer Rich Walke and Blake Wulfe and Ted Xiao and Jonathan Heewon Yang and Arefeh Yavary and Tony Z. Zhao and Christopher Agia and Rohan Baijal and Mateo Guaman Castro and Daphne Chen and Qiuyu Chen and Trinity Chung and Jaimyn Drake and Ethan Paul Foster and Jensen Gao and David Antonio Herrera and Minho Heo and Kyle Hsu and Jiaheng Hu and Donovon Jackson and Charlotte Le and Yunshuang Li and Kevin Lin and Roy Lin and Zehan Ma and Abhiram Maddukuri and Suvir Mirchandani and Daniel Morton and Tony Nguyen and Abigail O'Neill and Rosario Scalise and Derick Seale and Victor Son and Stephen Tian and Emi Tran and Andrew E. Wang and Yilin Wu and Annie Xie and Jingyun Yang and Patrick Yin and Yunchu Zhang and Osbert Bastani and Glen Berseth and Jeannette Bohg and Ken Goldberg and Abhinav Gupta and Abhishek Gupta and Dinesh Jayaraman and Joseph J Lim and Jitendra Malik and Roberto Martín-Martín and Subramanian Ramamoorthy and Dorsa Sadigh and Shuran Song and Jiajun Wu and Michael C. Yip and Yuke Zhu and Thomas Kollar and Sergey Levine and Chelsea Finn},
    booktitle = {Proceedings of Robotics: Science and Systems},
    year    = {2024},
}

@article{sontakke2024roboclip,
  title={Roboclip: One demonstration is enough to learn robot policies},
  author={Sontakke, Sumedh and Zhang, Jesse and Arnold, S{\'e}b and Pertsch, Karl and B{\i}y{\i}k, Erdem and Sadigh, Dorsa and Finn, Chelsea and Itti, Laurent},
  journal={Advances in Neural Information Processing Systems},
  volume={36},
  year={2024}
}

@inproceedings{ma2023liv,
  title={Liv: Language-image representations and rewards for robotic control},
  author={Ma, Yecheng Jason and Kumar, Vikash and Zhang, Amy and Bastani, Osbert and Jayaraman, Dinesh},
  booktitle={International Conference on Machine Learning},
  pages={23301--23320},
  year={2023},
  organization={PMLR}
}

@article{du2023vision,
  title={Vision-language models as success detectors},
  author={Du, Yuqing and Konyushkova, Ksenia and Denil, Misha and Raju, Akhil and Landon, Jessica and Hill, Felix and de Freitas, Nando and Cabi, Serkan},
  journal={arXiv preprint arXiv:2303.07280},
  year={2023}
}

@article{brohan2022rt,
  title={Rt-1: Robotics transformer for real-world control at scale},
  author={Brohan, Anthony and Brown, Noah and Carbajal, Justice and Chebotar, Yevgen and Dabis, Joseph and Finn, Chelsea and Gopalakrishnan, Keerthana and Hausman, Karol and Herzog, Alex and Hsu, Jasmine and others},
  journal={arXiv preprint arXiv:2212.06817},
  year={2022}
}

@inproceedings{andrychowicz2017hindsight,
  title={Hindsight experience replay},
  author={Andrychowicz, Marcin and Wolski, Filip and Ray, Alex and Schneider, Jonas and Fong, Rachel and Welinder, Peter and McGrew, Bob and Tobin, Josh and Abbeel, Pieter and Zaremba, Wojciech},
  booktitle={NeurIPS},
  year={2017}
}

@misc{zhou2023train,
      title={Train Offline, Test Online: A Real Robot Learning Benchmark}, 
      author={Gaoyue Zhou and Victoria Dean and Mohan Kumar Srirama and Aravind Rajeswaran and Jyothish Pari and Kyle Hatch and Aryan Jain and Tianhe Yu and Pieter Abbeel and Lerrel Pinto and Chelsea Finn and Abhinav Gupta},
      year={2023},
      eprint={2306.00942},
      archivePrefix={arXiv},
      primaryClass={cs.RO}
}

@article{lynch2023interactive,
  title={Interactive language: Talking to robots in real time},
  author={Lynch, Corey and Wahid, Ayzaan and Tompson, Jonathan and Ding, Tianli and Betker, James and Baruch, Robert and Armstrong, Travis and Florence, Pete},
  journal={IEEE Robotics and Automation Letters},
  year={2023},
  publisher={IEEE}
}

@article{jiang2024dexmimicgen,
  title={DexMimicGen: Automated Data Generation for Bimanual Dexterous Manipulation via Imitation Learning},
  author={Jiang, Zhenyu and Xie, Yuqi and Lin, Kevin and Xu, Zhenjia and Wan, Weikang and Mandlekar, Ajay and Fan, Linxi and Zhu, Yuke},
  journal={arXiv preprint arXiv:2410.24185},
  year={2024}
}

@inproceedings{mandlekar2018roboturk,
  title={Roboturk: A crowdsourcing platform for robotic skill learning through imitation},
  author={Mandlekar, Ajay and Zhu, Yuke and Garg, Animesh and Booher, Jonathan and Spero, Max and Tung, Albert and Gao, Julian and Emmons, John and Gupta, Anchit and Orbay, Emre and others},
  booktitle={Conference on Robot Learning},
  pages={879--893},
  year={2018},
  organization={PMLR}
}

@inproceedings{fang2024rh20t,
  title={Rh20t: A comprehensive robotic dataset for learning diverse skills in one-shot},
  author={Fang, Hao-Shu and Fang, Hongjie and Tang, Zhenyu and Liu, Jirong and Wang, Chenxi and Wang, Junbo and Zhu, Haoyi and Lu, Cewu},
  booktitle={2024 IEEE International Conference on Robotics and Automation (ICRA)},
  pages={653--660},
  year={2024},
  organization={IEEE}
}

@inproceedings{robomimic2021,
  title={What Matters in Learning from Offline Human Demonstrations for Robot Manipulation},
  author={Ajay Mandlekar and Danfei Xu and Josiah Wong and Soroush Nasiriany and Chen Wang and Rohun Kulkarni and Li Fei-Fei and Silvio Savarese and Yuke Zhu and Roberto Mart\'{i}n-Mart\'{i}n},
  booktitle={arXiv preprint arXiv:2108.03298},
  year={2021}
}

@inproceedings{ma2022vip,
  title={VIP: Towards Universal Visual Reward and Representation via Value-Implicit Pre-Training},
  author={Ma, Yecheng Jason and Sodhani, Shagun and Jayaraman, Dinesh and Bastani, Osbert and Kumar, Vikash and Zhang, Amy},
  booktitle={The Eleventh International Conference on Learning Representations},
  year={2022}
}

@inproceedings{liu2022robot,
    title = {Robot Learning on the Job: Human-in-the-Loop Autonomy and Learning During Deployment},
    author = {Huihan Liu and Soroush Nasiriany and Lance Zhang and Zhiyao Bao and Yuke Zhu},
    booktitle = {Robotics: Science and Systems (RSS)},
    year = {2023}
}

@inproceedings{vogel_edan_2020,
        title = {EDAN - an EMG-Controlled Daily Assistant to Help People with Physical Disabilities},
        language = {en},
        booktitle = {2020 {IEEE}/{RSJ} {International} {Conference} on {Intelligent} {Robots} and {Systems} ({IROS})},
        author = {Vogel, Jörn and Hagengruber, Annette and Iskandar, Maged and Quere, Gabriel and Leipscher, Ulrike and Bustamante, Samuel and Dietrich, Alexander and Hoeppner, Hannes and Leidner, Daniel and Albu-Schäffer, Alin},
        year = {2020}
}

@inproceedings{saxena2023multiresolution,
title={Multi-Resolution Sensing for Real-Time Control with Vision-Language Models},
author={Saumya Saxena and Mohit Sharma and Oliver Kroemer},
booktitle={7th Annual Conference on Robot Learning},
year={2023},
url={https://openreview.net/forum?id=WuBv9-IGDUA}
}

@inproceedings{bharadhwaj2023roboagent,
  title={Roboagent: Generalization and efficiency in robot manipulation via semantic augmentations and action chunking},
  author={Bharadhwaj, Homanga and Vakil, Jay and Sharma, Mohit and Gupta, Abhinav and Tulsiani, Shubham and Kumar, Vikash},
  booktitle={2024 IEEE International Conference on Robotics and Automation (ICRA)},
  pages={4788--4795},
  year={2024},
  organization={IEEE}
}

@article{levine2018learning,
  title={Learning hand-eye coordination for robotic grasping with deep learning and large-scale data collection},
  author={Levine, Sergey and Pastor, Peter and Krizhevsky, Alex and Ibarz, Julian and Quillen, Deirdre},
  journal={The International journal of robotics research},
  volume={37},
  number={4-5},
  pages={421--436},
  year={2018},
  publisher={SAGE Publications Sage UK: London, England}
}

@article{sermanet2016unsupervised,
  title={Unsupervised perceptual rewards for imitation learning},
  author={Sermanet, Pierre and Xu, Kelvin and Levine, Sergey},
  journal={arXiv preprint arXiv:1612.06699},
  year={2016}
}

@article{chen2021learning,
  title={Learning Generalizable Robotic Reward Functions from" In-The-Wild" Human Videos},
  author={Chen, Annie S and Nair, Suraj and Finn, Chelsea},
  journal={RSS},
  year={2021}
}

@article{levine2016end,
  title={End-to-end training of deep visuomotor policies},
  author={Levine, Sergey and Finn, Chelsea and Darrell, Trevor and Abbeel, Pieter},
  journal={The Journal of Machine Learning Research},
  volume={17},
  number={1},
  pages={1334--1373},
  year={2016},
}

@article{bu2025agibot,
  title={Agibot world colosseo: A large-scale manipulation platform for scalable and intelligent embodied systems},
  author={Bu, Qingwen and Cai, Jisong and Chen, Li and Cui, Xiuqi and Ding, Yan and Feng, Siyuan and Gao, Shenyuan and He, Xindong and Hu, Xuan and Huang, Xu and others},
  journal={arXiv preprint arXiv:2503.06669},
  year={2025}
}

@article{myers2023active,
  title={Active reward learning from online preferences},
  author={Myers, Vivek and B{\i}y{\i}k, Erdem and Sadigh, Dorsa},
  journal={arXiv preprint arXiv:2302.13507},
  year={2023}
}

@article{lee2020learning,
  title={Learning quadrupedal locomotion over challenging terrain},
  author={Lee, Joonho and Hwangbo, Jemin and Wellhausen, Lorenz and Koltun, Vladlen and Hutter, Marco},
  journal={Science robotics},
  volume={5},
  number={47},
  pages={eabc5986},
  year={2020},
  publisher={American Association for the Advancement of Science}
}

@article{wagenmaker2025steering,
  title={Steering Your Diffusion Policy with Latent Space Reinforcement Learning},
  author={Wagenmaker, Andrew and Nakamoto, Mitsuhiko and Zhang, Yunchu and Park, Seohong and Yagoub, Waleed and Nagabandi, Anusha and Gupta, Abhishek and Levine, Sergey},
  journal={arXiv preprint arXiv:2506.15799},
  year={2025}
}

@article{venuto2024code,
  title={Code as reward: Empowering reinforcement learning with vlms},
  author={Venuto, David and Islam, Sami Nur and Klissarov, Martin and Precup, Doina and Yang, Sherry and Anand, Ankit},
  journal={arXiv preprint arXiv:2402.04764},
  year={2024}
}

@article{baumli2023vision,
  title={Vision-language models as a source of rewards},
  author={Baumli, Kate and Baveja, Satinder and Behbahani, Feryal and Chan, Harris and Comanici, Gheorghe and Flennerhag, Sebastian and Gazeau, Maxime and Holsheimer, Kristian and Horgan, Dan and Laskin, Michael and others},
  journal={arXiv preprint arXiv:2312.09187},
  year={2023}
}

@article{rocamonde2023vision,
  title={Vision-language models are zero-shot reward models for reinforcement learning},
  author={Rocamonde, Juan and Montesinos, Victoriano and Nava, Elvis and Perez, Ethan and Lindner, David},
  journal={arXiv preprint arXiv:2310.12921},
  year={2023}
}

@misc{
  yang2023robot,
  title={Robot Fine-Tuning Made Easy: Pre-Training Rewards and Policies for Autonomous Real-World Reinforcement Learning},
  author={Jingyun Yang and Max Sobol Mark and Brandon Vu and Archit Sharma and Jeannette Bohg and Chelsea Finn},
  year={2023},
  eprint={2310.15145},
  archivePrefix={arXiv},
  primaryClass={cs.RO}
}

@inproceedings{shao2020concept,

 title={Concept2Robot: Learning Manipulation Concepts from Instructions and Human Demonstrations},

 author={Shao, Lin and Migimatsu, Toki and Zhang, Qiang and Yang, Karen and Bohg, Jeannette},

 booktitle={Proceedings of Robotics: Science and Systems (RSS)},

 year={2020},}
\bibliographystyle{iclr2026_conference}

\newpage
\appendix

\section{RoboReward Dataset}

\subsection{Dataset Sources}

\begin{table}[!htbp]
\caption{The various dataset sources comprising the RoboReward data mixture and benchmark. The resulting corpus contains \numtotalexamples{} examples in total: \numtrainexamples{} train, \numvalexamples{} validation, and \numtestexamples{} test.}
\label{tab:datasets}
\begin{center}
\setlength{\tabcolsep}{4pt}
\tiny
\begin{tabular}{p{2cm} p{2cm} p{5.4cm} p{1cm} r r r p{2.6cm}}
\multicolumn{1}{c}{\bf Name} &
\multicolumn{1}{c}{\bf Embodiment} &
\multicolumn{1}{c}{\bf Description} &
\multicolumn{1}{c}{\bf Perspective} &
\multicolumn{1}{c}{\bf Train} &
\multicolumn{1}{c}{\bf Val} &
\multicolumn{1}{c}{\bf Test} &
\multicolumn{1}{c}{\bf Citation}
\\ \midrule \\
RoboArena & DROID (Franka-based) & Distributed real-world evaluation episodes with per-episode progress scores and pairwise preferences. & Mix & 7337 & 1000 & 1000 & \cite{atreya2025roboarena} \\ 
Berkeley Bridge & WidowX & The robot interacts with household environments including kitchens, sinks, and tabletops. Skills include object rearrangement, sweeping, stacking, folding, and opening/closing doors and drawers. & Exocentric & 3826 & 496 & 100 & \cite{walke2023bridgedata} \\ 
Freiburg Franka Play & Franka & The robot interacts with toy blocks, it pick and places them, stacks them, unstacks them, opens drawers, sliding doors and turns on LED lights by pushing buttons. & Egocentric & 2856 & 352 & 91 & \cite{rosete2022tacorl,mees23hulc2} \\ 
USC Jaco Play & Jaco 2 & The robot performs pick-place tasks in a tabletop toy kitchen environment. & Exocentric & 2428 & 417 & 100 & \cite{dass2023jacoplay} \\ 
Roboturk & Sawyer & Sawyer robots flattens laundry, builds towers from bowls and searches objects. & Exocentric & 1463 & 0 & 97 & \cite{mandlekar2019scaling} \\ 
NYU VINN & Hello Stretch & The robot opens cabinet doors for a variety of cabinets. & Egocentric & 2369 & 0 & 0 & \cite{pari2021surprising} \\ 
Austin VIOLA & Franka & The robot performs various household-like tasks, such as setting up the table, or making coffee using a coffee machine. & Exocentric & 167 & 239 & 60 & \cite{zhu2022viola} \\ 
Berkeley Autolab UR5 & UR5 & The data consists of 4 robot manipulation tasks: simple pick-and-place of a stuffed animal between containers, sweeping a cloth, stacking cups, and a more difficult pick-and-place of a bottle that requires precise grasp and 6 DOF rotation. & Egocentric & 2388 & 430 & 100 & \cite{BerkeleyUR5Website} \\ 
TOTO & Franka & The TOTO Benchmark Dataset contains trajectories of two tasks: scooping and pouring. For scooping, the objective is to scoop material from a bowl into the spoon. For pouring, the goal is to pour some material into a target cup on the table. & Exocentric & 2986 & 0 & 0 & \cite{zhou2023train} \\ 
NYU ROT & xArm & The robot arm performs diverse manipulation tasks on a tabletop such an box opening, cup stacking, and pouring, among others. & Exocentric & 35 & 8 & 0 & \cite{haldar2023watch} \\ 
Stanford HYDRA & Franka & The robot performs the following tasks in corresponding environment: making a cup of coffee using the keurig machine; making a toast using the oven; sorting dishes onto the dish rack. & Exocentric & 507 & 203 & 91 & \cite{belkhale2023hydra} \\ 
Austin BUDS & Franka & The robot is trying to solve a long-horizon kitchen task by picking up pot, placing the pot in a plate, and push them together using a picked-up tool. & Exocentric & 127 & 0 & 0 & \cite{zhu2022bottom} \\ 
UCSD Kitchen & xArm & The dataset offers a comprehensive set of real-world robotic interactions, involving natural language instructions and complex manipulations with kitchen objects. & Exocentric & 393 & 122 & 95 & \cite{ucsd_kitchens} \\ 
UCSD Pick Place & xArm & The robot performs pick and place tasks in table top and kitchen scenes. The dataset contains a variety of visual variations. & Exocentric & 2384 & 0 & 100 & \cite{Feng2023Finetuning} \\ 
Austin Sirius & Franka & The dataset comprises two tasks, kcup and gear. The kcup task requires opening the kcup holder, inserting the kcup into the holder, and closing the holder. The gear task requires inserting the blue gear onto the right peg, followed by inserting the smaller red gear. & Exocentric & 1355 & 0 & 87 & \cite{liu2022robot} \\ 
Tokyo PR2 Fridge Opening & PR2 & PR2 opening/closing fridge and related appliance interactions. & Exocentric & 357 & 0 & 0 & \cite{oh2023pr2utokyodatasets} \\ 
Tokyo PR2 Tabletop Manipulation & PR2 & Reaching, grasping, placing on PR2 across varied objects and scenes. & Exocentric & 293 & 325 & 73 & \cite{oh2023pr2utokyodatasets} \\ 
UTokyo xArm PickPlace & xArm & The robot picks up a white plate, and then places it on the red plate. & Egocentric & 301 & 0 & 0 & \cite{matsushima2023weblab} \\ 
UTokyo xArm Bimanual & Dual xArms & The robots reach a towel on the table. They also unfold a wrinkled towel. & Egocentric & 161 & 0 & 71 & \cite{matsushima2023weblab} \\ 
Berkeley MVP & xArm & Basic motor control tasks (reach, push, pick) on table top and toy environments (toy kitchen, toy fridge). & Egocentric & 1218 & 228 & 71 & \cite{Radosavovic2022} \\ 
Berkeley RPT & Franka & Picking, stacking, destacking, and bin picking with variations in objects. & Egocentric & 1069 & 364 & 86 & \cite{Radosavovic2023} \\ 
KAIST Nonprehensile Objects & Franka & The robot performs various non-prehensile manipulation tasks in a tabletop environment. It translates and reorients diverse real-world and 3d-printed objects to a target 6 dof pose. & Exocentric & 406 & 162 & 53 & \cite{salhotra2022dmfd} \\ 
LSMO & Cobotta & The robot avoids obstacle on the table and reaches the target object. & Exocentric & 97 & 0 & 71 &  \\ 
CMU Franka Pick-Insert & Franka & The robot tries to pick up different-shaped objects placed in front of it. It also tries to insert particular objects into a cylindrical peg. & Exocentric & 1193 & 374 & 83 & \cite{saxena2023multiresolution} \\ 
Berkeley Fanuc Manipulation & Fanuc & A Fanuc robot performs various manipulation tasks. For example, it opens drawers, picks up objects, closes doors, closes computers, and pushes objects to desired locations. & Exocentric & 860 & 287 & 91 & \cite{Radosavovic2023} \\ 
CMU Play Fusion & Franka & The robot plays with 3 complex scenes: a grill with many cooking objects like toaster, pan, etc. It has to pick, open, place, close. It has to set a table, move plates, cups, utensils. And it has to place dishes in the sink, dishwasher, hand cups etc.  & Exocentric & 1269 & 358 & 92 & \cite{lynch2023interactive} \\ 
DROID & Franka & Various household manipulation tasks & Exocentric & 3071 & 378 & 100 & \cite{khazatsky2024droid} \\ 
RT-1 Robot Action & Google Robot & Robot picks, places and moves 17 objects from the google micro kitchens. & Exocentric & 3988 & 461 & 100 & \cite{brohan2022rt} \\ 
DLR Wheelchair Shared Control & DLR EDAN & The robot grasps a set of different objects in a table top and a shelf. & Exocentric & 168 & 28 & 19 & \cite{vogel_edan_2020} \\ 
\end{tabular}
\end{center}
\end{table}

\subsection{Data Cleaning and Augmentation Details} \label{app:cleaning}

For a successful episode $e=(v,t,r)$ with $r\!=\!5$, we construct additional training triples in two ways: (i) counterfactual commands $\tilde t$ paired with calibrated labels $\tilde r\in\{1,2,3,4\}$ for the same video $v$, and (ii) clipped videos $\tilde v$ paired with labels $\tilde r\in\{1,2,3,4\}$ for the original command $t$. For counterfactual relabeling, we use a multi-stage generation pipeline: (1) GPT-5 mini performs video analysis, (2) GPT-5 mini produces a plan for distinct failure modes that enforce a strict ordering $1<2<3<4<5$, (3) Qwen3 generates one imperative command per score in sequence, and (4) GPT-5 mini validates the resulting set and rejects it if it is inconsistent, triggering regeneration. For negative clipping, we generate a small ladder of clipped rollouts per episode and validate the resulting score assignments in the same way. Overall, this augmentation converts success videos into a balanced ladder of outcomes without fabricating videos and expands our training corpus.

\subsubsection{Prompt Rewrite (Invariant Clean-up)}\label{app:prompt_rewrite}

\paragraph{Model.} \texttt{Qwen3-4B-Instruct-2507} (text-only).

\paragraph{Purpose.} Correct grammar/spelling while preserving semantics (e.g., fix \emph{“palce dishes in the dish rack”} to \emph{“place the dishes in the dish rack”}).

\paragraph{Prompt.}
\begin{Verbatim}
Rewrite the following task description to correct grammar and spelling only.
Fix capitalization errors by beginning a sentence or phrase with a capital letter.
Do not change meaning.
Task description: {TASK}
Return only the corrected text.
\end{Verbatim}

\subsubsection{Negative Example Generation}\label{app:neg_example_gen}

\paragraph{Models.}
We use \texttt{gpt-5-mini-2025-08-07} for video analysis and planning, and \texttt{Qwen3-4B-Instruct-2507} for command generation.

\paragraph{Rubric (end-of-episode).}
\begin{Verbatim}
Rubric for end-of-episode progress (judge only the final state without time limits):
1 - No Success: Final state shows no goal-relevant change for the command.
2 - Minimal Progress: Final state shows a small but insufficient change toward the goal.
3 - Partial Completion: The final state shows good progress toward the goal but violates more than one requirement or a major requirement.
4 - Near Completion: Final state is correct in region and intent but misses a single minor requirement.
5 - Perfect Completion: Final state satisfies all requirements.
\end{Verbatim}

\paragraph{Video Analysis (VL).}
We sample the rollout video at 1 FPS and provide the resulting image frames (including the final frame) along with the original task description.
\begin{Verbatim}
You are analyzing a video of a robot performing a short-horizon manipulation task.
Describe the starting scene and objects visible. Be sure to mention objects in the task description.
Describe thoroughly object positions, locations and their relations to each other and to the robot for the starting scene.
Use the ROBOT BASE perspective for left/right; if the base is not visible, use the CAMERA perspective.
Then describe, step by step, what the robot does from start to end, focusing on the final state.
Be concrete and factual. Do not invent objects that are not visible.
Pay special attention to the last frame, especially when describing the final state.
Task description: {ORIGINAL_TASK}
Output sections:
1) Scene and objects
2) Robot actions step by step
3) Final state summary including the final state of the robot and relevant objects given the task description.
\end{Verbatim}

\paragraph{Planning (Text).}
\begin{Verbatim}
Plan carefully and step by step.
Goal: Think step by step to design distinct failure modes and concrete ideas for new task commands for scores 1, 2, 3, and 4, so that 1 < 2 < 3 < 4 < 5, where 5 is the original task fully satisfied by the video.
Judge only the final state and ignore time limits.
Use only visible objects and relations grounded in the video analysis.
Each score must correspond to a strictly closer final state to success than the previous one.
For reward 1, the robot performed no relevant action towards the goal (e.g., handled a completely different object than the one in the task description). Assign a distinct failure mode to each of 2, 3, and 4.
The ideas must not be entailed by or easier than the original task. Justify each idea's score in parentheses.
Critical: ideas for different scores must NOT be paraphrases or trivial rewrites of each other.
Each idea must NOT already be satisfied by the starting scene described above; require a visible change from the initial state.
Use the initial-state description explicitly to avoid proposing goals that are already true at the initial state.
For lower scores 2, 3, and 4 specifically, do NOT use control-action phrases similar to 'release the gripper' or 'let go of the gripper' or 'not in contact with the robot'.
For scores 3 and 4, refer to the SAME primary target object/entity using the same name as in the original task description. Do not switch to a different instance (e.g., 'cabinet' -> 'upper cabinet door').
Match the diction, writing style and complexity of the original task description.
Original task (score 5): {ORIGINAL_TASK}
Video analysis:
{VIDEO_ANALYSIS}
Rubric:
{RUBRIC}

Clarification for scoring 3 vs 4 (MINOR vs MAJOR requirements):
- Treat a requirement as MINOR when the PRIMARY object identity and PRIMARY spatial relation to the reference object are satisfied, but an AUXILIARY constraint is missed.
- Examples of AUXILIARY constraints that can be MINOR: remaining on the same support surface (e.g., keeping a cloth directly under the object), releasing/holding the gripper at the end, small orientation or placement tolerances, or cosmetic positioning details that do not change the core relation.
- Therefore, if the target object is correctly placed relative to the specified reference and all core elements are correct, but an auxiliary constraint like keeping the original cloth under the object is missed, prefer score 4 (Near Completion) rather than 3 (Partial Completion).
- Use score 3 only when multiple constraints are missed or when the missed constraint is CENTRAL to the task identity (e.g., primary spatial relation is incorrect).
Produce the following sections in order:
1) Reasoning - step-by-step analysis of the scene and of what final-state properties define success for the original task 5.
2) Separation plan - explain how to construct 1, 2, 3, and 4 so that 1 < 2 < 3 < 4 < 5 using only visible, concrete constraints.
3) Ideas for new task commands - propose highly focused candidate commands for each of scores 1, 2, 3, and 4. Justify each with a one-line reason for why it receives that score.
4) Monotonicity check - for each adjacent pair (1<2), (2<3), (3<4), and (4<5), justify why the later command is strictly closer to success under the rubric.
5) Final set of suggested commands - list the best final commands given the reasoning and separation plan for scores 1, 2, 3, 4, and 5 (remember 5 is the original task). Output only the commands with their scores, one per line.
\end{Verbatim}

\paragraph{Command Generation (Text, one score at a time).}

\begin{Verbatim}
Using the final suggested commands, generate a single imperative task command (one line) for the SAME video such that:
- Its correct evaluation under the rubric would be score {K} for the final state shown in the video.
- The command is stricter or different from the original so that the same video does not fully satisfy it if score is below 5.
- It is not entailed by the original task and is not an easier subset.
- Do not mention or reference the original task for the new task description.
- It uses only visible objects and relations described in the video analysis.
- Use plain ASCII characters only.
- Keep under 25 words.
- Start with a verb to make it an imperative command and do not mention the score or any meta-instructions in the command.
Original task (score 5): {ORIGINAL_TASK}
Rubric:
{RUBRIC}
Final suggested commands:
{PLAN_TEXT}
Keep in mind the following previously generated commands for lower scores and avoid duplication or contradiction while maintaining strict ordering (lower scores farther from success than higher scores):
{HISTORY}
Now output only the command for score {K} on a single line with no quotes and no extra commentary.
\end{Verbatim}

\subsubsection{Verification}\label{app:verification}

\paragraph{Model.} \texttt{gpt-5-mini-2025-08-07}.

\paragraph{Purpose.} Verify each example by checking (i) the task description is coherent and grounded in the video, and (ii) the provided discrete reward score matches the rubric. We keep only examples that pass this check.

\paragraph{Inputs.} We sample each rollout video at 1 FPS and pass the resulting image frames (including the true final frame) along with the task description and the provided score.

\paragraph{Prompt.}
\begin{Verbatim}
You are a strict annotation validator for short-horizon ROBOT TASK videos.

INPUTS
- TASK DESCRIPTION: {TASK_DESCRIPTION}
- PROVIDED SCORE: {PROVIDED_SCORE}  (must be one of 1,2,3,4,5)

YOUR JOB (single-pass):
1) Watch the video.
2) Using the rubric below, evaluate the robot's actions during the task and the final state, and determine whether the PROVIDED SCORE is correct.
3) Check that the TASK DESCRIPTION is coherent and refers to the correct items/actions visible in the video.
4) Decision basis and rejection criteria:
   - If the TASK DESCRIPTION is incoherent with the video (e.g., wrong primary object identity, inconsistent relation, key details not visible, or rule conflicts), you MUST output ANSWER: FALSE.
   - Otherwise, output ANSWER: TRUE if the PROVIDED SCORE matches the rubric for the video; output ANSWER: FALSE if it does not match.

RUBRIC (end-of-episode, discrete progress):
1 - No Success: Final state shows no goal-relevant change for the command (e.g., wrong object was handled)
2 - Minimal Progress: Final state shows a small but insufficient change toward the goal.
3 - Partial Completion: The final state shows good progress toward the goal but violates more than one requirement or a major requirement.
4 - Near Completion: The final state is correct in region and intent but misses a single minor requirement.
5 - Perfect Completion: Fully satisfies the task as described; ALL requirements met.

MINOR VS MAJOR REQUIREMENTS (clarification for 3 vs 4):
- Treat a requirement as MINOR when the PRIMARY object identity and PRIMARY spatial relation to the reference object are satisfied, but an AUXILIARY constraint is missed.
- Examples of AUXILIARY constraints that can be MINOR: remaining on the same support surface (e.g., keeping a cloth directly under the object), holding/releasing the gripper at the end, small orientation or placement tolerances, or cosmetic positioning details that do not change the core relation.
- Therefore, if the target object is correctly placed relative to the specified reference and all core elements are correct, but an auxiliary constraint like keeping the original cloth under the object is missed, prefer score 4 (Near Completion) rather than 3 (Partial Completion).
- Use score 3 only when multiple constraints are missed or when the missed constraint is CENTRAL to the task identity (e.g., wrong object identity or the primary spatial relation is incorrect).

PERSPECTIVE & COMMON PITFALLS (apply strictly):
- Left/Right: Use the ROBOT BASE perspective. If the base is not visible, use the CAMERA perspective.
- Object naming: REJECT if the description mislabels key items.
- Consistency rules:
  * If the primary target object/entity name in the TASK DESCRIPTION does not match the object named in the video, treat as mismatch -> REJECT. Do NOT treat category synonyms or toy/real variants as equivalent (e.g., "sandwich slice" vs "toy cheese").
  * End location naming: Allow common surface synonyms (e.g., towel/cloth/napkin/rag) as equivalent IF color and relative placement match, and there is no conflicting evidence. If color or relative placement clearly differ, REJECT.
- Irrelevant objects: Ignore objects not needed to perform the task (do NOT reject just for extra/unused items).
- Move vs slide: Do not conflate sliding on a surface with picking up/placing if the description depends on this distinction.
- Color ambiguity: If color terms feel ambiguous BUT the intended item/action is still unambiguous for executing the task, do NOT reject for color alone.
- Visibility: If key details to judge success are not visible or the video is too unclear, REJECT.

TASK SANITY CHECKS (MUST REJECT if any apply):
- Task string is empty, placeholder, or nonsense and not a valid task description/command (e.g., "No image loaded in this hit").
- Task does not reference any visible objects/relations in the video or cannot be grounded.

OUTPUT FORMAT (STRICT)
1) First, write your reasoning (few sentences) that clearly states whether the PROVIDED SCORE matches the rubric and why.
2) Then on a new line write exactly ONE of the following without any formatting and nothing else:
   ANSWER: TRUE
   or
   ANSWER: FALSE
\end{Verbatim}

\subsection{Human Verification for RoboRewardBench} \label{app:human_verification}

\begin{figure}[H]
  \centering
  \includegraphics[width=\textwidth]{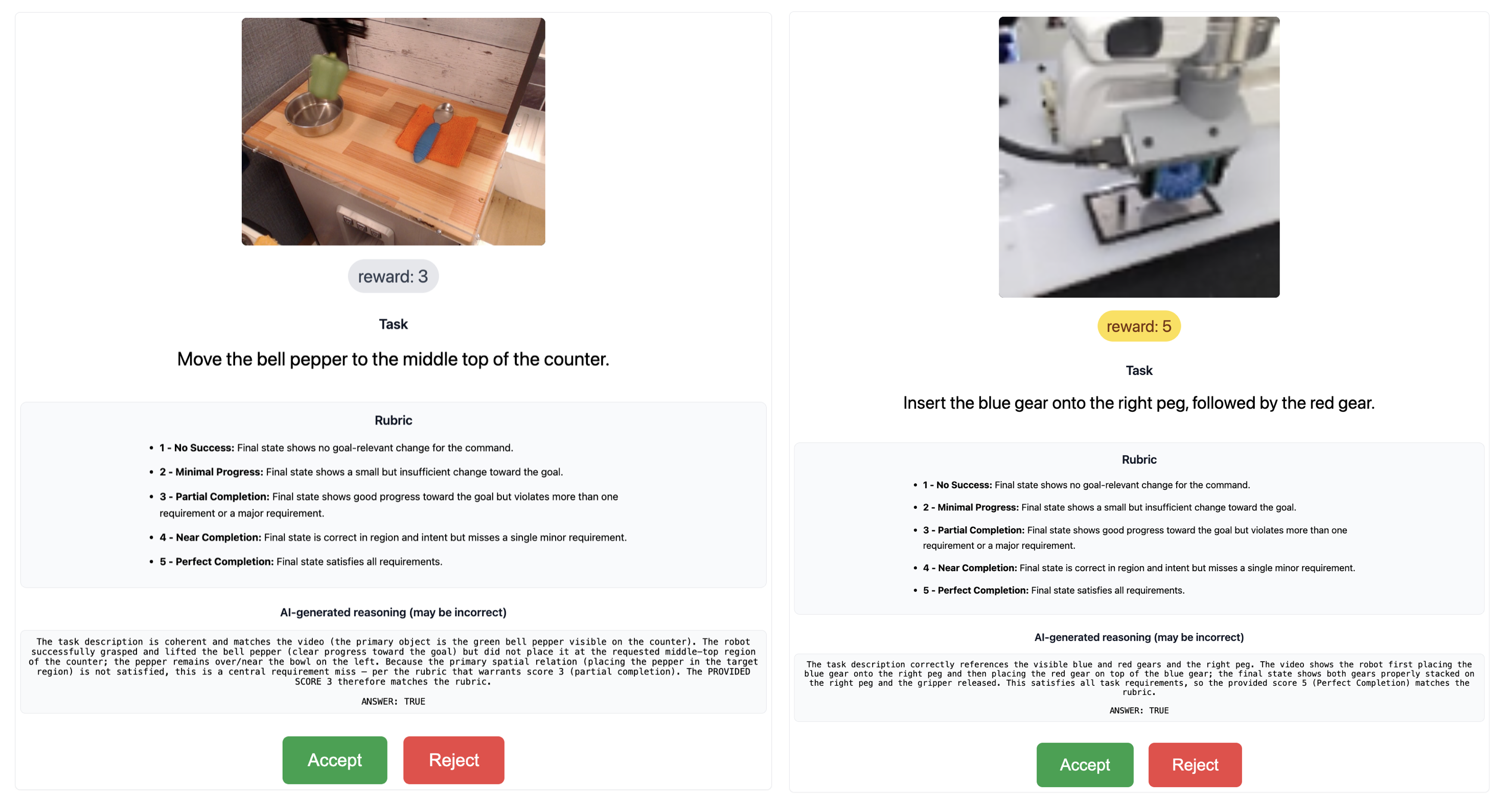}
  \caption{Annotation UI used for human verification. Annotators watch the rollout video and are shown the task text, the provided reward label, the rubric, and the GPT-5 mini verification rationale produced by our automated validation step. They then accept or reject the example based on whether the reward label is justified by the video under the rubric.}
  \label{fig:annotation_ui}
\end{figure}

To construct a higher-trust evaluation suite, we additionally perform \emph{human} verification for the test split. Concretely, each example is reviewed by one human annotator, who is asked to confirm that the end-of-episode reward label is justified under our rubric, given the rollout video and task description (see example at \Cref{fig:annotation_ui}). When a mismatch is found, we discard the example. We then subsample from the remaining verified examples to form a clean evaluation set. The resulting human-verified test split contains \numtestexamples{} examples, which we refer to as \textbf{RoboRewardBench}.

\section{Experimental details}\label{app:more_expt_details}

\subsection{Robomimic Experiments}

For all \texttt{Robomimic} experiments, we run DSRL-NA with the hyperparameters as specified for each environment in \cite{wagenmaker2025steering}. For completeness, we include these in \Cref{tab:dsrl_online_hyperparams,tab:dsrl_robomimic_hyperparams}.

\begin{table}[H]
\caption{
\footnotesize
\textbf{Common DSRL hyperparameters for \texttt{Robomimic} experiments.}
}
\vspace{5pt}
\label{tab:dsrl_online_hyperparams}
\begin{center}
\scalebox{0.9}
{
\begin{tabular}{ll}
    \toprule
    \textbf{Hyperparameter} & \textbf{Value} \\
    \midrule
    Learning rate & $0.0003$ \\
    Batch size & $256$ \\
    Activation & Tanh \\
    Target entropy & $0$ \\
    Target update rate ($\tau$) & $0.005$ \\
    Number of actor and critic layers & $3$ \\
    Number of critics & $2$ \\
    Number of environments & $4$ \\
    \bottomrule
\end{tabular}
}
\end{center}
\end{table}

\begin{table}[H]
\caption{
\footnotesize
\textbf{Hyperparameters for DSRL \texttt{Robomimic} experiments.}
}
\vspace{5pt}
\label{tab:dsrl_robomimic_hyperparams}
\begin{center}
\scalebox{0.9}
{
\begin{tabular}{llll}
    \toprule
    \textbf{Hyperparameter} & \texttt{Lift} & \texttt{Can} & \texttt{Square} \\
    \midrule
    Action chunk size & $4$ & $4$ & $4$  \\
Hidden size & $2048$  & $2048$ & $2048$   \\
Gradient steps per update & $30$ & $20$ & $20$   \\
$Q^{\mathcal{W}}$ update steps & $10$ & $10$ & $10$  \\
Discount factor & $0.99$ & $0.99$ & $0.999$  \\
Action magnitude & $1.5$ & $1.5$ & $1.5$   \\
Initial steps & $24000$ & $24000$ & $32000$ \\
Train denoising steps & $20$ & $20$ & $100$  \\
Inference denoising steps & $8$ & $8$ & $8$  \\
    \bottomrule
\end{tabular}
}
\end{center}
\end{table}

\subsection{Benchmarking with RoboRewardBench}

\begin{table}[H]
\footnotesize
\caption{Vision--language models evaluated on \textbf{RoboRewardBench} and their overall results. Rows are ordered by overall group-wise mean absolute error (MAE; lower is better). \textit{Limited} denotes models available only via a restricted API at the time of evaluation, for which little public technical information is available (e.g., parameter count).}
\label{tab:models}
\centering
\resizebox{\linewidth}{!}{
\begin{tabular}{r l l c c c c}
\toprule
\textbf{Rank} & \textbf{Model} & \textbf{Creator} & \textbf{Parameters} & \textbf{Access} & \textbf{MAE} & \textbf{Reference} \\
\midrule
1  & RoboReward (8B)                         & \textbf{Ours} & 8B   & Open    & 0.665 & This work \\
2  & GPT-5 mini (2025-08-07)                 & OpenAI              & --   & Limited & 0.691 & \cite{openai_gpt5_system_card} \\
3  & GPT-5 (2025-08-07)                      & OpenAI              & --   & Limited & 0.811 & \cite{openai_gpt5_system_card} \\
4  & RoboReward (4B)                         &  \textbf{Ours}           & 4B   & Open    & 0.845 & This work \\
5  & Gemini 3 Pro (Preview)                  & Google DeepMind     & --   & Limited & 0.851 & \cite{google_gemini3_pro_model_card} \\
6  & GPT-5.2 (2025-12-11)                    & OpenAI              & --   & Limited & 0.887 & \cite{openai_gpt52_system_card} \\
7  & Qwen3-VL Instruct (8B)                  & Alibaba & 8B   & Open    & 0.892 & \cite{qwen3vl_tech_report} \\
8  & GPT-5.1 (2025-11-13)                    & OpenAI              & --   & Limited & 0.901 & \cite{openai_gpt51_system_card} \\
9  & Gemini 2.5 Pro                          & Google DeepMind            & --   & Limited & 0.902 & \cite{google_gemini25_tech_report} \\
10 & Qwen3-VL Instruct (30B)                 & Alibaba & 30B  & Open    & 0.903 & \cite{qwen3vl_tech_report} \\
11 & Gemini Robotics-ER 1.5                  & Google DeepMind     & --   & Limited & 0.906 & \cite{google_gemini_robotics15_tech_report} \\
12 & Gemini 3 Flash (Preview)                & Google DeepMind               & --   & Limited & 0.917 & \cite{google_gemini3_flash_model_card} \\
13 & Gemini 2.5 Flash                        & Google DeepMind              & --   & Limited & 0.943 & \cite{google_gemini25_tech_report} \\
14 & Gemini 2.5 Flash-Lite                   & Google DeepMind             & --   & Limited & 0.990 & \cite{google_gemini25_flashlite_model_card} \\
15 & Qwen2.5-VL Instruct (72B)               & Alibaba & 72B  & Open    & 0.991 & \cite{bai2025qwen25vltechnicalreport} \\
16 & Qwen3-VL Instruct (4B)                  & Alibaba & 4B   & Open    & 1.032 & \cite{qwen3vl_tech_report} \\
17 & Qwen2.5-VL Instruct (32B)               & Alibaba & 32B  & Open    & 1.137 & \cite{bai2025qwen25vltechnicalreport} \\
18 & Qwen2.5-VL Instruct (7B)                & Alibaba & 7B   & Open    & 1.172 & \cite{bai2025qwen25vltechnicalreport} \\
19 & Llama 4 Maverick Instruct               & Meta                & --   & Open    & 1.271 & \cite{meta_llama4_model_card} \\
20 & GPT-5 nano (2025-08-07)                 & OpenAI              & --   & Limited & 1.295 & \cite{openai_gpt5_system_card} \\
21 & Llama 4 Scout Instruct                  & Meta                & --   & Open    & 1.485 & \cite{meta_llama4_model_card} \\
22 & Qwen2.5-VL Instruct (3B)                & Alibaba & 3B   & Open    & 1.607 & \cite{bai2025qwen25vltechnicalreport} \\
\bottomrule
\end{tabular}
}
\vspace{2mm}
\end{table}

\subsection{The Pitfalls of State-of-the-Art VLMs as Reward Models}\label{app:vlm_pitfalls}

Our benchmarking results (\Cref{sec:rrbench_results}) and real-world RL experiments (\Cref{sec:real_world_rl}) show that frontier VLMs can act as reward models, but they still make simple mistakes that are easy for humans to catch. In this section, we provide qualitative examples that help explain these failure modes.

We focus on \textbf{Gemini Robotics-ER 1.5} because it is a frontier model trained on robotics data and designed for embodied reasoning and progress estimation \citep{google_gemini_robotics_er}. Despite these targeted capabilities, it still produces incorrect progress scores on real robot rollouts, including during RL training.

\paragraph{Gemini Robotics-ER 1.5 makes high-impact reward mistakes.}

\begin{figure}[t] \centering \includegraphics[width=\linewidth]{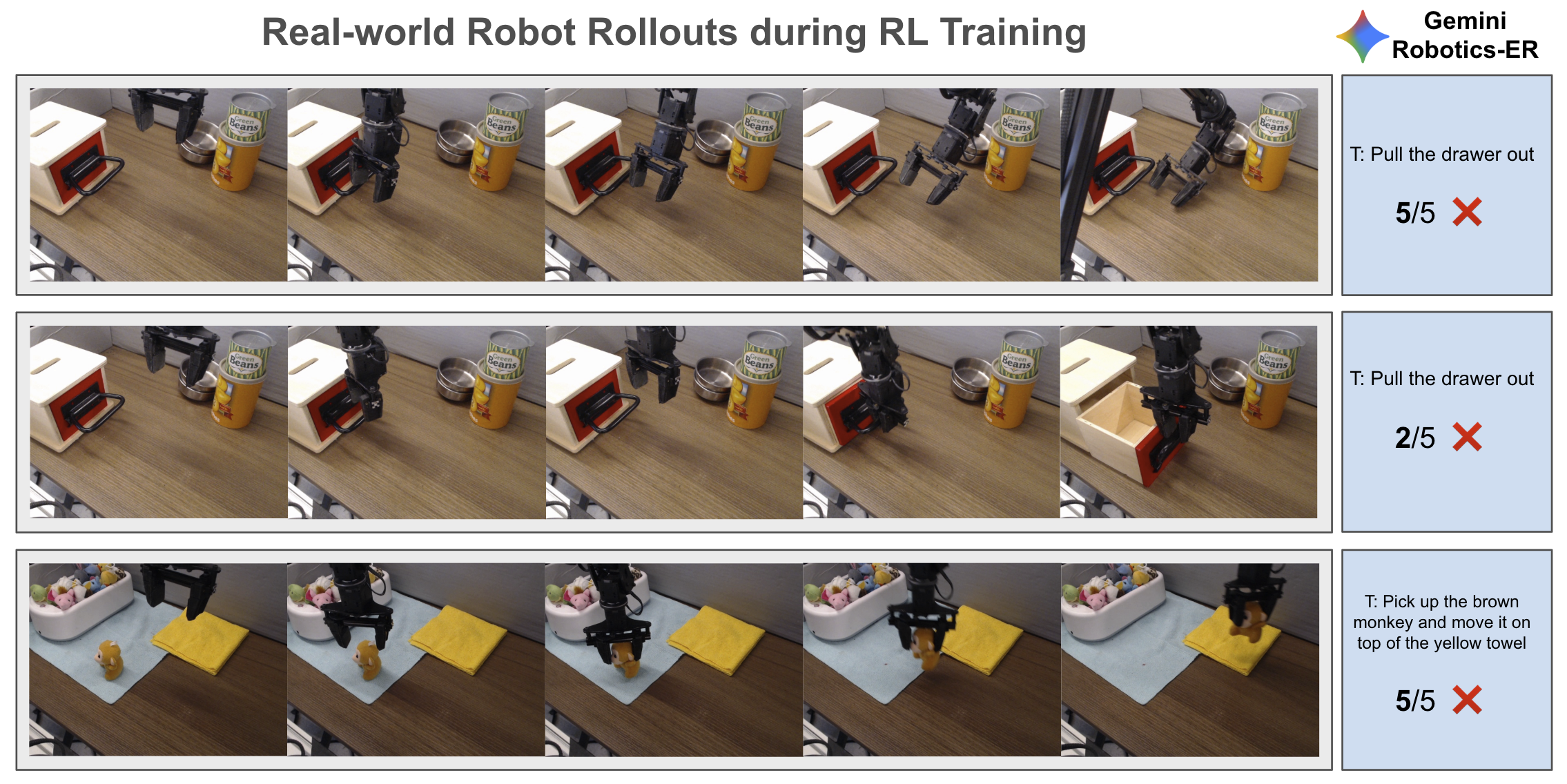} 
    \caption{\textbf{Failure cases of Gemini Robotics-ER 1.5 as a reward model on real robot rollouts.}
    Each row shows a rollout (frames left) and the model’s end-of-episode progress score given the video rollout and task (right).
    \textbf{Top:} \textit{Pull the drawer out.} The robot collapses to the right of the handle and the drawer remains closed, but the model predicts \textbf{5/5} (false positive).
    \textbf{Middle:} \textit{Pull the drawer out.} The robot successfully pulls the drawer open, but the model predicts \textbf{2/5} (false negative).
    \textbf{Bottom:} \textit{Pick up the brown monkey and move it on top of the yellow towel.} The final frame shows the monkey held above the towel (obviously hovering rather than clearly placed), yet the model predicts \textbf{5/5}, showing a lack of simple spatial reasoning.}
    \label{fig:gemini_failures}
\end{figure}

\Cref{fig:gemini_failures} shows three representative errors.
For row 1, the task is \textit{pull the drawer out}. The robot does not successfully grasp the handle, the drawer remains closed, and the gripper falls over to the right. Nevertheless, Gemini Robotics-ER assigns a score of 5/5.
For row 2, the robot this time succeeds in pulling the drawer open, but Gemini Robotics-ER assigns only a 2/5 instead of a perfect completion score of 5, leading to a false negative.
Finally, for row 3, the task is \textit{pick up the brown monkey and move it on top of the yellow towel}. 
The final frame shows the monkey held above the towel (not clearly placed), yet the model deems the trajectory ends with a perfect completion, assigning a reward of 5/5.
These kinds of false negatives and positives are particularly problematic when assigning rewards during RL: over-rewarding failures can reinforce the wrong behavior, and under-rewarding successes can weaken the learning signal.

To better understand these mis-scored rewards and verify whether the model is hallucinating key events, we prompt Gemini Robotics-ER 1.5 \emph{``Describe the video in great detail.''} with default hyperparameters:

\begin{tcolorbox}[breakable,title={Row 1: ``Pull the drawer out'' (model assigns 5/5 on failure)},
                 colback=gray!4, colframe=gray!60, boxrule=0.6pt, arc=2pt]
\begin{Verbatim}
Based on the video frames, here is a detailed description of the scene and action:

The video begins with a static shot of a wooden table surface. On the left side of the frame sits a small, light-colored wooden box. The front face of the box has a red panel with a black metal handle attached, featuring a looped design. In the background, to the right of the box, are several items: a stack of containers consisting of a yellow canister at the bottom and a green can labeled "Green Beans" on top, and a small, nested metal bowl next to them.

Starting at 00:01, a black robotic arm, specifically its end-effector or gripper, moves into the frame from the top right. The gripper is articulated with two fingers. It descends towards the wooden box on the left.

From 00:03 to 00:04, the robot arm positions the open gripper directly in front of the black handle on the box. The fingers of the gripper are carefully aligned with the handle.

By 00:05, the gripper's fingers close around the black handle, securing its grasp on the box.

Finally, from 00:06 to 00:07, the robot arm pulls back slightly, dragging the wooden box with it by the handle. A part of the robot's vertical structure briefly appears on the far left edge of the frame. The action demonstrates the robot's ability to identify and manipulate objects using the handle.
\end{Verbatim}
\end{tcolorbox}

Gemini Robotics-ER 1.5 correctly describes the scene: it correctly identifies the main objects (the wooden box with a red front panel and black handle, the stacked canisters labeled ``Green Beans,'' and the metal bowls) and provides a plausible high-level intent of the action (the arm approaching the handle). However, the description incorrectly asserts a precise grasp (``fingers close around the handle'') and a successful pull (``dragging the wooden box'') even though the rollout does not clearly show secure contact or drawer motion. This suggests a failure mode where the model interpolates a typical manipulation narrative from partial visual evidence, rather than tracking the true temporal progression and fine-grained spatial state needed for reward assignment.

\begin{tcolorbox}[breakable,title={Row 2: ``Pull the drawer out'' (model assigns 2/5 on success)},
                 colback=gray!4, colframe=gray!60, boxrule=0.6pt, arc=2pt]
\begin{Verbatim}[fontsize=\footnotesize]
Based on the sequence of images, here is a detailed description of the scene and action:

The video shows a high-angle view of a robot arm interacting with objects on a wooden tabletop. On the left side of the frame is a light-colored wooden box with a prominent black D-shaped handle attached to a red rectangular plate on its front face. Behind this box, and to the right, are several other items: a stack of two metal bowls, a large yellow cylindrical container, and a smaller green can labeled "Green Beans" sitting on top of the yellow container.

At the beginning of the clip (00:00), a black, two-fingered robot gripper is visible in the upper center part of the frame, poised in the air above the wooden surface.

As the video progresses, the robot arm moves steadily from right to left (00:01-00:02). The gripper arm adjusts its position, lowering itself and rotating to align with the black handle on the wooden box.

By the final frame (00:03), the robot arm's gripper has moved significantly closer and is positioned directly in front of the handle, appearing ready to grasp it. The movement captures the robot arm in the process of targeting and approaching the handle on the wooden box.
\end{Verbatim}
\end{tcolorbox}

The description again captures the static scene well: it correctly identifies the drawer box, the D-shaped handle on the red plate, and the surrounding objects (bowls, yellow container, ``Green Beans'' can), and it accurately describes the gripper’s approach and alignment behavior. The failure is that it stops short of narrating the decisive task outcome. In the rollout, the drawer is actually pulled open, but the model’s description ends at ``appearing ready to grasp,'' suggesting it under-tracks late-stage temporal events and state changes.

\begin{tcolorbox}[breakable,title={Row 3: ``Pick up the brown monkey and move it on top of the yellow towel'' (model assigns 5/5 without a clear placement)},
                 colback=gray!4, colframe=gray!60, boxrule=0.6pt, arc=2pt]
\begin{Verbatim}
Based on the video frames, a robotic gripper performs a sequence of actions involving a plush toy.

1.  **Initial State (00:00):** The scene features a small brown monkey plush toy on a light blue mat. To the right of the mat, on a wooden surface, is a folded yellow cloth. A robotic gripper is positioned above the objects.
2.  **Grasping Action (00:01 - 00:04):** The robotic gripper moves down towards the brown monkey plush toy. The gripper's claws open and descend around the toy. The claws then close, grasping the monkey and lifting it from the blue mat.
3.  **Relocation (00:05 - 00:07):** The gripper, holding the monkey, lifts it higher and then moves horizontally to the right, carrying the toy over the wooden surface towards the yellow cloth.
4.  **Placement (00:08):** The gripper positions the brown monkey directly above the folded yellow cloth and lowers it, placing the toy onto the cloth.
\end{Verbatim}
\end{tcolorbox}

The model states a completed placement (“placing the toy onto the cloth”), but the rollout does not clearly show the monkey resting on the towel. Instead, the final frame is ambiguous and consistent with the monkey still being held above the towel. The description, therefore, hallucinates a key success event (the release/placement), which leads to an incorrect  \textbf{5/5} score.

\paragraph{Takeaway.}
These examples suggest that even frontier VLMs trained on robotics data can understand the scene at a high level while still failing on the fine-grained spatial and temporal details that reward modeling depends on (e.g., whether a grasp is secure, whether a drawer actually opens, or whether an object is truly placed versus hovering). In practice, both RoboReward 8B and Gemini Robotics-ER 1.5 often produce reasonable rewards, but these kinds of false rewards are enough to contribute to errors on RoboRewardBench (\Cref{sec:rrbench_results}) and create noticeable gaps to oracle human rewards in real-world RL (\Cref{sec:real_world_rl}).

\subsection{Real-World RL Experiment}\label{app:real_world_rl}

DSRL finetunes the behavior of a base diffusion policy by modifying the input noise given to its denoising process
For the base diffusion policy, we utilize the diffusion transformer architecture due to \cite{dasari2025ingredients}. We pretrain the diffusion policy on the BridgeData V2 dataset and use goal image conditioning to specify the task. See \Cref{tab:dit_params} for the hyperparameters used for the diffusion policy. 

For DSRL, we utilize the DSRL-SAC variant described in \cite{wagenmaker2025steering}. We follow the procedure used for the WidowX experiments in \cite{wagenmaker2025steering} exactly. In particular, we first roll out the base diffusion policy 20 times to warm-start the replay buffer, and then start running RL training. RL hyperparameters are given in \Cref{tab:widowx_dsrl_params} (all reward models use the same DSRL hyperparameters).

\begin{table}
\caption{
\footnotesize
\textbf{Hyperparameters for base diffusion policy in WidowX experiments.}
}
\label{tab:dit_params}
\vspace{5pt}
\begin{center}
\begin{tabular}{ll}
    \toprule
    \textbf{Hyperparameter} & Value \\
    \midrule
Batch size &  $2048$ \\
Learning rate & $0.0003$ \\
Training steps & $100000$ \\
LR scheduler & cosine \\
Warmup steps & $2000$ \\
Action chunk size & $1$ \\
Train denoising steps & $100$ \\
Inference denoising steps & $8$ \\
Image encoder & ResNet-34 \\
Hidden size & $256$ \\
Number of Heads & $1$ \\
Number of Layers & $3$ \\
Feedforward dimension & $512$ \\
    \bottomrule
\end{tabular}
\end{center}
\end{table}

\begin{table}
\caption{
\footnotesize
\textbf{Hyperparameters for DSRL WidowX experiments.}
}
\label{tab:widowx_dsrl_params}
\vspace{5pt}
\begin{center}
\begin{tabular}{llll}
    \toprule
    \textbf{Hyperparameter} & Value \\
    \midrule
Hidden size & $1024$     \\
Gradient steps per update  & $30$    \\
Discount factor & $0.97$    \\
Action magnitude & $1.5$    \\
    \bottomrule
\end{tabular}
\end{center}
\end{table}

\begin{landscape}
\begin{table}[p]
\centering
\tiny
\begin{tabular}{p{2.4cm}p{0.8cm}p{0.48cm}p{0.48cm}p{0.48cm}p{0.48cm}p{0.48cm}p{0.48cm}p{0.48cm}p{0.48cm}p{0.48cm}p{0.48cm}p{0.48cm}p{0.48cm}p{0.48cm}p{0.48cm}p{0.48cm}p{0.48cm}p{0.48cm}p{0.48cm}p{0.48cm}p{0.48cm}p{0.48cm}p{0.48cm}p{0.48cm}}
\toprule
\parbox[t]{2.4cm}{Model} & \parbox[t]{0.8cm}{\textbf{Mean Absolute Error \\(↓ \\better)}} & \parbox[t]{0.48cm}{Robo Arena} & \parbox[t]{0.48cm}{Austin Sirius} & \parbox[t]{0.48cm}{Berkeley Autolab UR5} & \parbox[t]{0.48cm}{Berkeley Fanuc Manipulation} & \parbox[t]{0.48cm}{Berkeley MVP} & \parbox[t]{0.48cm}{Berkeley RPT} & \parbox[t]{0.48cm}{Berkeley Bridge} & \parbox[t]{0.48cm}{CMU Play Fusion} & \parbox[t]{0.48cm}{DLR Wheelchari Shared Control} & \parbox[t]{0.48cm}{DROID} & \parbox[t]{0.48cm}{RT-1 Robot Action} & \parbox[t]{0.48cm}{CMU Franka Pick-Insert} & \parbox[t]{0.48cm}{USC Jaco Play} & \parbox[t]{0.48cm}{KAIST Nonprehensile Objects} & \parbox[t]{0.48cm}{Roboturk} & \parbox[t]{0.48cm}{Stanford HYDRA} & \parbox[t]{0.48cm}{Freiburg Franka Play} & \parbox[t]{0.48cm}{LSMO} & \parbox[t]{0.48cm}{UCSD Kitchen} & \parbox[t]{0.48cm}{UCSD Pick Place} & \parbox[t]{0.48cm}{Tokyo PR2 Tabletop Manipulation} & \parbox[t]{0.48cm}{UTokyo xArm Bimanual} & \parbox[t]{0.48cm}{Austin VIOLA} \\
\midrule
RoboReward 8B \textbf{(ours)} & \textbf{0.665} & \textbf{0.768} & 0.701 & \textbf{0.340} & \textbf{0.538} & 0.577 & 0.674 & 0.770 & \textbf{0.587} & 0.474 & 0.670 & 0.830 & 0.807 & \textbf{0.720} & \textbf{0.396} & \textbf{0.485} & 0.890 & \textbf{0.473} & 0.493 & 0.947 & \textbf{0.100} & 0.452 & 1.394 & 1.200 \\
GPT-5 mini (2025-08-07) & 0.691 & 0.862 & 0.379 & 0.640 & 0.769 & \textbf{0.408} & 0.581 & \textbf{0.530} & 1.141 & \textbf{0.158} & \textbf{0.640} & \textbf{0.630} & 0.566 & 1.090 & 0.774 & 0.505 & 0.835 & 1.077 & 0.676 & 0.832 & 0.480 & \textbf{0.411} & 1.592 & 0.317 \\
GPT-5 (2025-08-07) & 0.811 & 1.028 & 0.310 & 0.830 & 0.934 & 0.845 & 1.174 & 0.790 & 0.989 & 0.211 & 0.760 & 0.830 & 1.048 & 0.830 & 0.868 & 0.629 & 0.670 & 1.110 & 0.648 & 1.063 & 0.510 & 0.918 & 1.366 & 0.283 \\
RoboReward 4B \textbf{(ours)} & 0.845 & 0.806 & 0.897 & 0.700 & 0.681 & 0.887 & 0.721 & 0.910 & 0.663 & 0.579 & 0.880 & 0.890 & 0.843 & 0.950 & 0.698 & 0.969 & 1.176 & 0.846 & 0.535 & \textbf{0.705} & 0.330 & 0.562 & 1.282 & 1.933 \\
Gemini 3 Pro (Preview) & 0.851 & 1.234 & 0.379 & 0.690 & 0.824 & 1.394 & \textbf{0.395} & 0.750 & 0.630 & 0.316 & 0.820 & 1.040 & 0.795 & 0.890 & 1.491 & 0.990 & 0.846 & 0.989 & 0.972 & 1.242 & 0.370 & 0.438 & 1.789 & 0.283 \\
GPT-5.2 (2025-12-11) & 0.887 & 0.973 & 0.759 & 0.780 & 1.165 & 0.521 & 1.407 & 0.940 & 1.185 & 0.263 & 0.910 & 0.940 & \textbf{0.506} & 0.910 & 1.057 & 0.557 & 0.780 & 1.308 & 0.986 & 1.105 & 0.500 & 0.822 & 1.718 & 0.317 \\
Qwen3-VL Instruct (8B) & 0.892 & 0.847 & 0.713 & 0.420 & 0.890 & 0.817 & 0.919 & 0.980 & 1.163 & 0.474 & 0.890 & 1.050 & 1.000 & 0.950 & 0.755 & 0.794 & 1.000 & 1.385 & 1.423 & 0.895 & 0.540 & 0.822 & 1.085 & 0.700 \\
GPT-5.1 (2025-11-13) & 0.901 & 1.075 & 0.575 & 0.960 & 0.923 & 0.676 & 1.012 & 1.190 & 1.000 & 0.368 & 1.160 & 1.000 & 0.855 & 0.940 & 0.906 & 0.619 & 1.066 & 1.242 & 0.761 & 1.316 & 0.570 & 0.589 & 1.437 & 0.483 \\
Gemini 2.5 Pro & 0.902 & 0.936 & 0.333 & 0.920 & 0.736 & 1.296 & 0.709 & 1.170 & 0.804 & 0.842 & 0.920 & 1.200 & 0.687 & 0.890 & 1.113 & 1.309 & \textbf{0.495} & 1.187 & 0.915 & 1.316 & 0.580 & 0.466 & 1.746 & 0.167 \\
Qwen3-VL Instruct (30B) & 0.903 & 0.993 & 0.632 & 0.780 & 1.121 & 0.465 & 0.988 & 1.160 & 1.337 & 0.474 & 1.070 & 1.040 & 0.590 & 1.010 & 0.849 & 0.897 & 1.088 & 1.099 & 0.746 & 1.032 & 0.610 & 0.973 & 1.423 & 0.383 \\
Gemini Robotics-ER 1.5 & 0.906 & 1.002 & 0.448 & 0.730 & 0.791 & 1.394 & 0.442 & 1.000 & 0.891 & 0.895 & 0.760 & 0.980 & \textbf{0.506} & 0.980 & 1.208 & 1.165 & 0.846 & 1.176 & 0.521 & 1.326 & 0.750 & 0.658 & 2.070 & 0.300 \\
Gemini 3 Flash (Preview) & 0.917 & 1.058 & \textbf{0.276} & 0.550 & 0.736 & 1.352 & 0.430 & 0.990 & 0.707 & 1.053 & 0.770 & 1.130 & 0.819 & 1.100 & 1.509 & 1.134 & 0.659 & 1.088 & 0.549 & 1.716 & 0.950 & 0.767 & 1.648 & \textbf{0.100} \\
Gemini 2.5 Flash & 0.943 & 1.190 & 0.655 & 0.850 & 0.791 & 1.254 & 0.826 & 1.080 & 0.783 & 0.526 & 0.830 & 1.430 & \textbf{0.506} & 0.990 & 1.358 & 0.979 & 0.956 & 1.341 & \textbf{0.465} & 1.516 & 0.760 & 0.671 & 1.676 & 0.250 \\
Gemini 2.5 Flash-Lite & 0.990 & 1.136 & 0.609 & 1.110 & 1.154 & 0.958 & 1.221 & 1.070 & 1.065 & 0.368 & 1.080 & 1.130 & 1.012 & 0.960 & 0.849 & 1.021 & 1.044 & 1.198 & 1.197 & 1.316 & 0.440 & 0.863 & 1.465 & 0.500 \\
Qwen2.5-VL Instruct (72B) & 0.991 & 0.848 & 0.575 & 1.100 & 1.110 & 0.437 & 1.209 & 1.180 & 0.739 & 1.105 & 0.920 & 1.270 & 0.711 & 1.190 & 0.887 & 0.876 & 1.110 & 1.264 & 1.141 & 1.200 & 1.060 & 0.452 & 2.141 & 0.267 \\
Qwen3-VL Instruct (4B) & 1.032 & 0.929 & 1.115 & 0.680 & 1.187 & 1.155 & 0.930 & 1.140 & 1.011 & 0.895 & 1.210 & 1.230 & 0.807 & 1.250 & 1.094 & 1.113 & 1.176 & 1.165 & 1.099 & 1.063 & 0.920 & 0.959 & \textbf{1.000} & 0.600 \\
Qwen2.5-VL Instruct (32B) & 1.137 & 1.064 & 0.782 & 1.510 & 1.099 & 0.775 & 1.302 & 1.150 & 1.033 & 1.316 & 1.010 & 1.220 & 0.867 & 1.130 & 1.151 & 1.000 & 1.626 & 1.308 & 1.479 & 1.347 & 0.910 & 0.521 & 1.704 & 0.850 \\
Qwen2.5-VL Instruct (7B) & 1.172 & 1.035 & 1.253 & 1.190 & 1.165 & 0.789 & 1.523 & 1.120 & 1.141 & 1.421 & 0.970 & 0.970 & 1.217 & 1.200 & 1.453 & 1.175 & 1.132 & 1.242 & 1.014 & 1.211 & 0.840 & 0.973 & 1.310 & 1.617 \\
Llama 4 Maverick Instruct & 1.271 & 1.552 & 1.023 & 1.260 & 1.000 & 0.789 & 1.651 & 1.130 & 1.283 & 1.684 & 1.490 & 1.020 & 1.735 & 1.250 & 0.849 & 0.990 & 1.319 & 1.099 & 1.606 & 1.358 & 0.840 & 0.699 & 1.465 & 2.133 \\
GPT-5 nano (2025-08-07) & 1.295 & 1.290 & 1.345 & 1.000 & 1.319 & 1.085 & 1.616 & 1.310 & 1.511 & 1.368 & 1.340 & 1.350 & 1.253 & 1.440 & 1.000 & 0.814 & 1.341 & 1.659 & 1.366 & 1.232 & 1.300 & 1.411 & 1.521 & 0.917 \\
Llama 4 Scout Instruct & 1.485 & 1.830 & 1.241 & 1.010 & 1.505 & 1.070 & 1.419 & 1.440 & 1.370 & 1.211 & 1.340 & 1.500 & 1.843 & 1.480 & 1.340 & 1.351 & 2.165 & 1.418 & 1.746 & 1.568 & 1.030 & 1.123 & 1.606 & 2.550 \\
Qwen2.5-VL Instruct (3B) & 1.607 & 1.443 & 1.632 & 1.360 & 1.593 & 1.718 & 1.837 & 1.430 & 1.413 & 1.895 & 1.430 & 1.380 & 1.530 & 1.420 & 1.566 & 1.320 & 1.769 & 1.714 & 1.648 & 1.453 & 1.430 & 1.548 & 1.690 & 2.733 \\
\bottomrule
\end{tabular}
\caption{Full benchmarking results on RoboRewardBench. Models are ranked by mean absolute error (lower is better). The columns report absolute error for each dataset subset.}
\label{fig:all}
\end{table}
\end{landscape}

\end{document}